\documentclass[lettersize,journal]{IEEEtran}
\usepackage{amsmath}
\usepackage{amsfonts}
\usepackage{amssymb}
\usepackage{algorithm}
\usepackage{algpseudocode}
\usepackage{array}
\usepackage[caption=false,font=scriptsize,labelfont=sf,captionskip=0pt]{subfig}
\usepackage{textcomp}
\usepackage{stfloats}
\usepackage{url}
\usepackage{verbatim}
\usepackage{graphicx}
\usepackage{cite}
\usepackage{bm}
\usepackage{accents}
\usepackage{array}
\usepackage{multicol}
\usepackage{multirow}
\usepackage{eqparbox}
\usepackage{enumerate}
\usepackage{booktabs}
\usepackage{soul}
\usepackage{xcolor}
\usepackage{nccmath}
\usepackage{makecell}
\usepackage{amsthm}
\usepackage{diagbox}
\usepackage{hyperref}
\usepackage{etoolbox}

\newtheorem{theorem}{Theorem}

\newtheorem{remark}[theorem]{Remark}

\newcommand{\prob}{\mathrm{I\kern-0.15em P}}

\begin{document}
\title{A Unified Variational Imputation Framework for Electric Vehicle Charging Data\\ Using Retrieval-Augmented Language Model}

\author{Jinhao Li
~and~Hao Wang,~\IEEEmembership{Member~IEEE}
\thanks{This work was supported in part by the Australian Research Council (ARC) Discovery Early Career Researcher Award (DECRA) under Grant DE230100046. (Corresponding author: Hao Wang.)}
\thanks{J. Li and H. Wang are with the Department of Data Science and AI, Faculty of IT and Monash Energy Institute, Monash University, Melbourne, VIC 3800, Australia (e-mails: \{jinhao.li, hao.wang2\}@monash.edu).}
}



\maketitle
\thispagestyle{empty}

\begin{abstract}
The reliability of data-driven applications in electric vehicle (EV) infrastructure, such as charging demand forecasting, hinges on the availability of complete, high-quality charging data. However, real-world EV datasets are often plagued by missing records, and existing imputation methods are ill-equipped for the complex, multimodal context of charging data, often relying on a restrictive one-model-per-station paradigm that ignores valuable inter-station correlations. To address these gaps, we develop a novel \textit{PR}obabilistic variational imputation framework that leverages the power of large l\textit{A}nguage models and retr\textit{I}eval-augmented \textit{M}emory (PRAIM). PRAIM employs a pre-trained language model to encode heterogeneous data, spanning time-series demand, calendar features, and geospatial context, into a unified, semantically rich representation. This is dynamically fortified by retrieval-augmented memory that retrieves relevant examples from the entire charging network, enabling a single, unified imputation model empowered by variational neural architecture to overcome data sparsity. Extensive experiments on four public datasets demonstrate that PRAIM significantly outperforms established baselines in both imputation accuracy and its ability to preserve the original data's statistical distribution, leading to substantial improvements in downstream forecasting performance. 
\end{abstract}


\section{Introduction} \label{sec:intro}
The global transition towards electric mobility is accelerating, with electric vehicle (EV) sales already accounting for more than one in four cars sold worldwide in 2025 \cite{IEA2025}. The correspondingly expanding charging network generates a vast stream of data, chronicling every charging session's time, duration, and consumed energy. These data are the bedrock for numerous downstream applications, such as demand forecasting and grid load management \cite{Li_2024}. The efficacy of these data-driven tasks hinges on a fundamental prerequisite: the availability of complete and high-quality datasets. However, real-world EV charging data are often plagued by missing records \cite{fahim2025lstmcnn}, a less-noticed but critical challenge that significantly undermines its utility and necessitates effective data imputation.

\subsection{Motivation} \label{subsec:intro_motivation}
While leveraging EV charging data for downstream tasks has been extensively studied, the integrity of the underlying datasets has rarely been investigated. By aggregating raw charging sessions into daily demand, we observe that publicly accessible datasets have a considerable amount of missing daily records, which is discussed in detail in Section \ref{sec:dataset_overview}. The reason behind these missing values is often uncertain -- it could stem from system malfunctions such as communication errors, or it could reflect days with no genuine charging activity. Discerning between these two possibilities is non-trivial, yet accurately imputing these missing values offers significant advantages from two perspectives: 1) augmenting the original dataset, which can potentially improve the performance of downstream tasks and 2) recovering the true data distribution, enabling a more comprehensive analysis of real-world charging behaviors.

Approaches to data imputation can be broadly categorized as follows. Initially, simple statistical methods, such as mean or median value imputation \cite{majidpour2014meanzero}, provide straightforward solutions. However, their inability to capture complex data relationships leads to the adoption of machine learning (ML)-based methods \cite{wang2025survey,ben2022svd,xu2023svd,Demirhan2018kalman,vanBuuren2011mice,Stekhoven2011missforest}, such as K-nearest neighbors (KNN) \cite{wang2025survey}, which can model nonlinear, feature-wise interactions. Though effective, these methods often struggle with the temporal dependencies inherent in the datasets, further addressed by the emerging deep learning (DL)-based methods \cite{lee2020lstm,bansal2021_Transformer,chen2022graph,kong2023graph,miao2021GAN,shen2022GAN,ge2024GAN,kuppannagari2021autoencoder,liguori2023autoencoder,xiao2023diffusion,tashiro2021diffusion}.

However, general-purpose imputation methods are less sufficient to address the unique challenges of imputing EV charging data. The first is limited data availability. Most public datasets span only a few years, providing a restricted history that makes it difficult for data-hungry ML and DL models to achieve satisfactory accuracy. Though recent transfer or few-shot learning \cite{R2C1_1,R2C1_4} could alleviate general data scarcity, charging stations, in particular newly deployed or recently reconnected ones, often have short operational histories, long contiguous gaps, and burstiness (i.e., weeks of missing followed by weeks of data). Such unique data characteristics make parametric adaptation challenging.

A second, significant limitation is the need to train an isolated model for each station. This one-model-per-station paradigm fails to leverage informative inter-station correlations that are essential for understanding the contextual demand patterns, in particular within an urban charging network. While graph-based DL methods \cite{chen2022graph,kong2023graph} are expected to overcome this problem by considering neighboring stations, they introduce a critical vulnerability that is the need for complete data sequences from both the target station and its neighbors to generate training samples. However, given the considerable amount of missing data, this requirement is often untenable as shown in Appendix A, creating a dilemma where the models best suited for utilizing spatial context are the most crippled by missing data.

A further challenge lies in effectively leveraging the rich, multimodal context that governs EV charging demand patterns. Accurate imputation depends not only on data availability but also on a range of other contextual factors. For instance, calendar features, like the day of the week, define the primary rhythm of demand, while the geospatial context, revealed by nearby points of interest (PoIs) such as offices or shopping centers, dictates a station's functional role. The interaction between these modalities is paramount, e.g., demand at a station in a business district on a weekday could be significantly different from that on a weekend. However, most studies focus on the numerical time-series of charging demand, either ignoring the rich, unstructured context from geospatial and textual data or treating it as a simple categorical input (often encoded via one-hot encoding) \cite{ben2022svd,lee2020lstm}. Without effectively integrating such heterogeneous information and capturing their underlying interactions between data types, crucial predictive signals could be substantially lost, thereby limiting the imputation accuracy. Therefore, a powerful and unified feature engineering mechanism is needed to fuse these disparate data sources into a single, cohesive representation that characterizes their intricate relationships.

Given the identified challenges, we are motivated to develop a unified imputation framework, creating a single, universal model capable of imputing data for all charging stations by leveraging inter-station information and effectively encoding diverse, multimodal data sources.

\subsection{Literature Review} \label{subsec:intro_literature}
Data imputation spans a spectrum from simple statistical fillers to advanced learning-based models. At the low end, statistical methods replace missing entries with summaries of observed data via mean, median, zero imputation, or interpolation, which are computationally inexpensive but can distort the underlying distribution \cite{wang2025survey}. Classical ML methods seek greater accuracy by exploiting structure: KNN imputes from neighbors in feature space \cite{wang2025survey} and matrix factorization decomposes data to uncover latent patterns \cite{xu2023svd,ben2022svd}; additional approaches include Kalman filters for time series \cite{Demirhan2018kalman}, multivariate imputation by chained equations \cite{vanBuuren2011mice}, and MissForest, which leverages random forests \cite{Stekhoven2011missforest}. More recently, deep learning has become prominent for modeling complex, nonlinear patterns, with two broad lines: 1) prediction-based imputation that treats imputation as forecasting via long short-term memory (LSTM) \cite{lee2020lstm}, transformer \cite{bansal2021_Transformer}, and graph neural network \cite{chen2022graph,kong2023graph}, and 2) reconstruction-based imputation that learns the data distribution using generative models, including generative adversarial networks (GANs) \cite{miao2021GAN,shen2022GAN,ge2024GAN}, denoising autoencoders \cite{kuppannagari2021autoencoder,liguori2023autoencoder}, and diffusion models \cite{xiao2023diffusion,tashiro2021diffusion}.

Despite the extensive research, related work specifically on EV charging data imputation remains scarce. The few existing studies have proposed simpler or adapted versions of the reviewed methods. For example, early work used median and zero imputation \cite{majidpour2014meanzero}. More recent studies have explored DL models, including LSTM \cite{lee2020lstm}, a combination of bidirectional-LSTM and convolutional neural networks \cite{fahim2025lstmcnn}, and GANs \cite{shen2022GAN,ge2024GAN}. However, as discussed in Section \ref{subsec:intro_motivation}, these adopted models cannot fully address the identified challenges. For example, statistical fillers assume short, isolated gaps and locally regular dynamics. When charging missingness arrives in long, contiguous blocks, these methods either bridge outages by inventing activity where none is likely or collapse rare peaks toward the mean. Also, prediction-based deep-learning models require continuous supervision to learn temporal dependencies, but atypical missingness removes exactly the sequences needed for training. Further, reconstruction-based methods like GAN rely on abundant, representative samples per station to learn meaningful latent representations. With short, ragged histories and shifting semantics across stations, learned representations skew toward average behavior and systematically under-represent rare yet operationally salient usage behaviors.

The reviewed studies highlight a clear gap for a more effective imputation framework for EV charging data.

\subsection{Main Work and Contributions} \label{subsec:intro_contributions}
To address the identified gaps, we develop a novel \textbf{pr}obabilistic variational imputation framework, termed \textit{PRAIM}, that synthesizes the remarkable encoding capability of large l\textbf{a}nguage models (LLMs) with a dynamic, retr\textbf{i}eval-based \textbf{m}emory. 

We first confront the challenge of multimodal data fusion by leveraging a pre-trained LLM, which acts as a universal interpreter (equipped with vast pre-trained knowledge) to translate the heterogeneous context for each data point, spanning time-series charging demand, calendar features, and geospatial PoIs, into a single, semantically rich embedding. Recent advancements of LLMs in spatiotemporal data analysis, such as UrbanGPT \cite{R2C2_urbangpt} and ST-LLM+ \cite{R2C2_stllm}, further support our use of LLM as a spatiotemporal context interpreter, helping recover missing demand patterns while preserving the underlying spatiotemporal structure. Further, to tackle data availability, the LLM-generated embedding is then dynamically fortified by a retrieval-augmented generation (RAG) mechanism that treats the entire training dataset \textit{across all stations} as an external knowledge base, retrieving the most relevant scenarios to inform the current imputation. The integration of RAG effectively leverages statistical strengths from other stations or time periods with similar contexts. More importantly, since the RAG enhances each data point with relevant context from the whole charging network, PRAIM is no longer confined to station-specific patterns and can be trained as a single, universal imputation engine for all stations, breaking from the restrictive one-model-per-station paradigm.

Finally, the RAG-augmented representation conditions a tailored variational sequence-to-sequence neural architecture to generate not only the most likely imputed values but also a measure of their uncertainty. Our contributions are as follows.
\begin{itemize}
    \item \textit{Unified LLM-Based Imputation Framework}: We use a pre-trained LLM as a unified context encoder to fuse heterogeneous inputs, including time-series demand, calendar, geospatial PoIs, into a semantically rich representation, overcoming a key limitation of prior imputation models that struggle to integrate such diverse data sources.
    
    \item \textit{RAG-Driven Contextual Enrichment}: We introduce RAG to treat the entire multi-station dataset as a dynamic knowledge base. By retrieving and fusing context from the most relevant samples, RAG enhances LLM-generated embeddings and is the crucial enabler for training a unified imputation engine.
    
    \item \textit{Variational Probabilistic Imputation for Uncertainty Quantification}: Moving beyond deterministic methods, PRAIM is built on a variational sequence-to-sequence framework that outputs a probabilistic distribution for each reconstructed value, providing a more realistic imputation with quantified uncertainty.
    
    \item \textit{Extensive Empirical Validation}: Using four real-world datasets, we demonstrate that PRAIM not only outperforms a wide array of benchmarks in imputation accuracy but also excels at preserving the underlying statistical distribution of the data. We further show that this high-fidelity imputation translates into significant performance gains for downstream forecasting models.
\end{itemize}

The remainder of this paper is organized as follows. Section \ref{sec:dataset_overview} provides an overview of used EV charging datasets with analysis of missing data impact. Section \ref{sec:data_prep} details the data preparation and feature engineering process. Section \ref{sec:model_formulation} presents the architecture of PRAIM. Section \ref{sec:model_train_eval} outlines the model training and evaluation procedures. Section \ref{sec:experiments} details experimental results. Finally, Section \ref{sec:conclusion} concludes the paper.

\section{EV Charging Dataset Overview} \label{sec:dataset_overview}
\begin{figure}[!t]
    \centering
    \includegraphics[width=.85\linewidth]{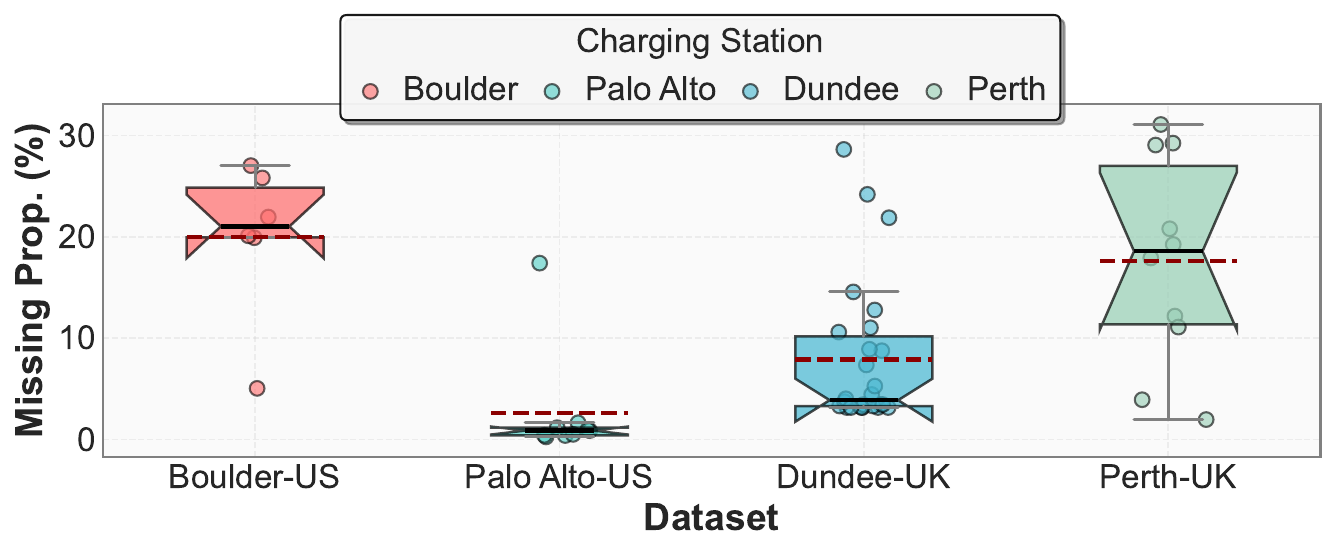}
    \caption{Distribution of missing daily record percentages in the four datasets. The dashed and solid lines indicate the mean and median, respectively.}
    \label{fig:missing_dist}
\end{figure}

Our study uses four publicly available, real-world datasets of EV charging sessions from Palo Alto (USA) \cite{dataCityofPaloAlto}, Dundee (UK) \cite{dataCityofDundee}, Boulder (USA) \cite{dataCityofBoulder}, and Perth (UK) \cite{dataCityofPerth}. To ensure a baseline level of data quality and exclude potentially abandoned or malfunctioning charging units, we filter out charging stations from all four datasets with more than $35\%$ missing daily records, i.e., the percentage of specific days that do not have recorded charging is above $35\%$.
\begin{itemize}
    \item \textit{Palo Alto} contains usage data from nine public charging stations in Palo Alto, California, with records spanning from July 2011 to December 2020. As shown in Fig. \ref{fig:missing_dist}, the dataset presents a median missing data proportion of less than $2\%$. 

    \item \textit{Dundee} originally includes 30 stations, sourced from the city of Dundee, Scotland. After our filtering process, 26 stations remain. The data covers the period from 2017 to 2018, with a moderate level of missing data--a median of approximately $8\%$ across the stations.

    \item \textit{Boulder} originally contains records from 27 charging stations, provided by the city of Boulder, Colorado. After applying the $35\%$ threshold, only six stations are retained, which highlights a significant data availability challenge. The data spans from 2018 to 2023.

    \item \textit{Perth} tracks usage for public charging points from Perth and Kinross Council, Scotland. Initially containing 13 stations, ten remain after filtering. The data spans from December 2016 to December 2019. Alongside the Boulder dataset, it is one of the more challenging datasets due to a high proportion of missing values.
\end{itemize}
Additionally, we also provide a theoretical analysis regarding the impact of missing data on the downstream forecasting in Appendix A.

\section{Data Preparation} \label{sec:data_prep}
This section details the transformation of the raw, event-based EV charging data into a structured, feature-rich format. Section \ref{subsec:data_structuring} structures the raw data by aggregating charging sessions into daily time-series, along with geospatial and calendar features. Section \ref{subsec:feature_engineering} details feature engineering on 1) generating feature embedding via LLM and 2) refining features through demand normalization and cyclical encoding.

\subsection{Data Structuring} \label{subsec:data_structuring}
The original dataset consists of a log of charging sessions, where each entry includes a station identifier, timestamp, duration, energy consumed, and station's geographic coordinates, denoted as $\mathbf{g}_i$, where $i$ represents the station index.

\subsubsection{Temporal Aggregation}
The event-based charging sessions are converted into a time series by discretizing the time domain into daily intervals. For a given station $i$ and day $t$, the daily charging demand, defined as $d_{t,i}$, is calculated by summing the energy from all sessions that occur at that station within the 24-hour period. In addition to the demand, we extract calendar attributes for each day, including the day of the week, day of the month, and month of the year, which are summarized into a vector denoted as $\mathbf{c}_t \in \mathbb{R}^3$.

We aggregate these sessions into daily resolution rather than finer temporal resolution (e.g., hourly) for three main reasons that matter to imputation: 1) it could align with the dominant weekly rhythm in EV usage, such as weekday/weekend cycles and holiday effects, 2) daily aggregation avoids bin-allocation heuristics of  (e.g., prorating partial sessions across sub-hourly bins) and suppresses telemetry noise and short outages, resulting in a cleaner target for imputation and for forecasting tasks conducted at day-to-week horizons, and 3) sub-daily charging is highly stochastic, which is dominated by uncontrollable human behaviors, including arrival/departure timing, dwell, opportunistic top-ups. As a result, minute/hourly signals could present large idiosyncratic variance that is hard to model and forecast reliably without additional real-time covariates.

\subsubsection{Geospatial Feature Extraction}
To capture the local environmental context that influences charging behaviors (e.g., workplace v.s. residential), for each station $i$, we query the OpenStreetMap database \cite{OpenStreetMap} to identify a set of nearby PoIs denoted as $\mathcal{POI}_i$, which includes all PoIs whose locations fall within a predefined distance from the station's location. 

\subsection{Feature Engineering} \label{subsec:feature_engineering}
Given the structured data, we create two feature streams as inputs into the PRAIM, including a semantically-rich, LLM-generated embedding and a set of carefully prepared features.

\subsubsection{LLM-Empowered Contextual Embedding Generation} \label{subsubsec:data_prep_llm_emb_generation}
As previously discussed in Section \ref{subsec:intro_motivation}, the contextual information of EV charging is highly heterogeneous, thereby requiring powerful embedding. Inspired by LLM's ability to process and reason over a combination of text, numbers, and symbols within a unified framework \cite{liangle2024}, we leverage a pre-trained LLM to fuse disparate data sources. Unlike specialized encoding models that excel at a single data type, LLMs, with vastly pre-trained knowledge, can interpret the semantic meaning and contextual relevance of all inputs simultaneously.

For station $i$ on day $t$, a detailed textual prompt is constructed by a prompt generation function $f(\cdot)$, which integrates six distinct types of information:
\begin{itemize}
    \item \textit{Station Identifier}: A categorical variable denoted as $v_i$;

    \item \textit{Daily Charging Demands:} The most recent $L$ days of charging demands, $\mathbf{d}_{t,i} = [d_{t-L+1, i}, \dots, d_{t,i}]$;

    \item \textit{Missing Data Indicator:} A binary vector indicating missing entries over the same $L$-day window, $\mathbf{m}_{t,i} = [m_{t-L+1, i}, \dots, m_{t,i}]$, with element $(\mathbf{m}_{t,i})_{t'} \in \{0, 1\}$, where $t'\in \{t-L+1, \cdots, t\}$. A value of one suggests that the demand on day $t'$ is missing;
    
    \item \textit{Calendar Features:} $\mathbf{c}_t$;

    \item \textit{Geospatial Coordinates:} $\mathbf{g}_i$;

    \item \textit{Points of Interest:} $\mathcal{POI}_i$.
\end{itemize}

\begin{figure}[!t]
    \centering
    \includegraphics[width=.85\linewidth]{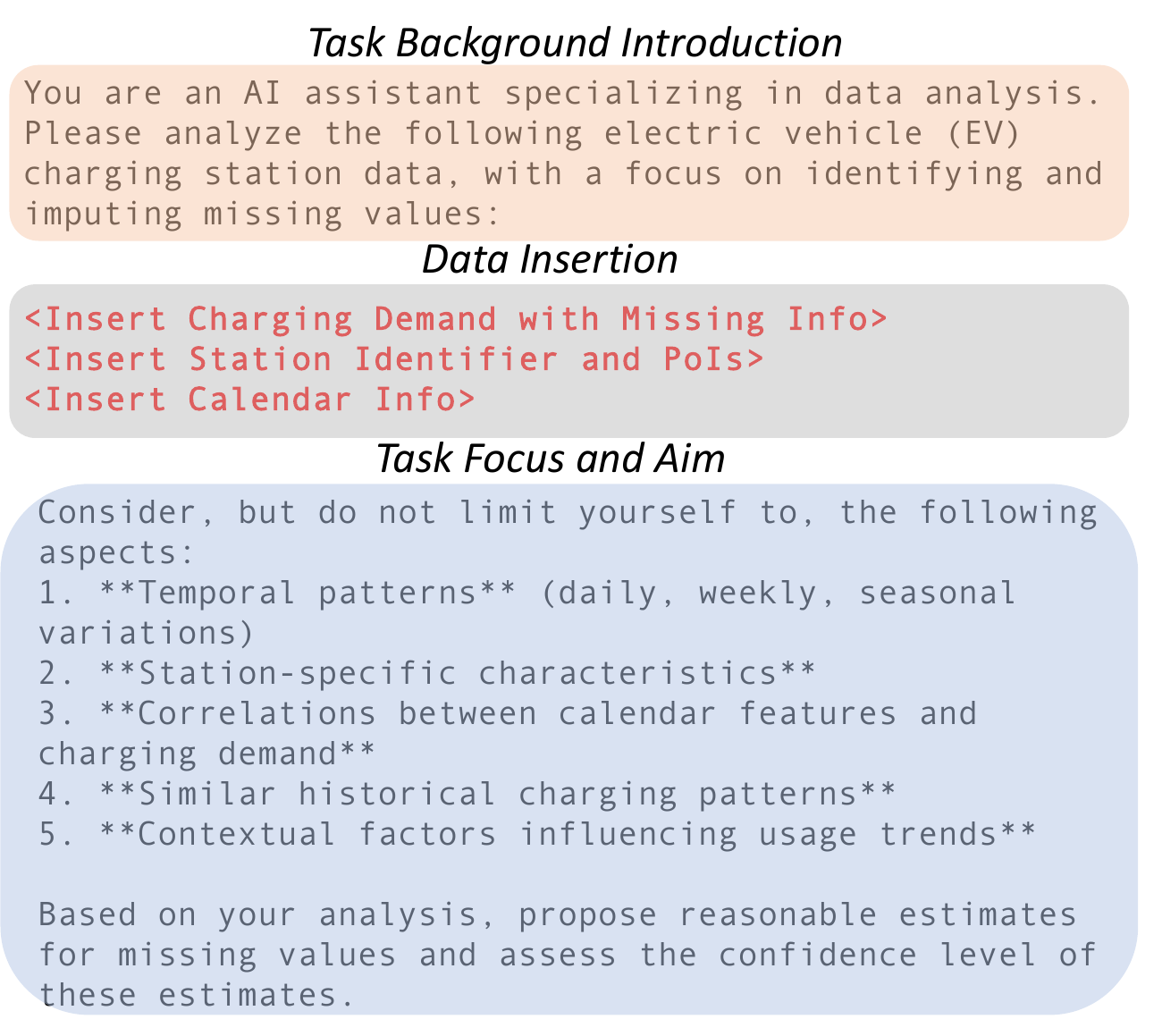}
    \caption{An example of prompt construction for EV charging station data.}
    \label{fig:example_prompt}
\end{figure}

An example of how these information streams are framed into a single prompt is depicted in Fig. \ref{fig:example_prompt}. LLM takes the prompt to produce a fixed-size embedding vector, denoted as $\boldsymbol{\omega}_{t,i} \in d_\mathrm{emb}$, where $d_\mathrm{emb}$ represents the embedding dimension. The embedding process can be formulated as
\begin{equation}
\label{eq:llm_embedding}
    \boldsymbol{\omega}_{t,i} = \mathcal{LLM}\left( f( v_i, \mathbf{d}_{t,i}, \mathbf{m}_{t,i}, \mathbf{c}_t, \mathbf{g}_i, \mathcal{POI}_i) \right).
\end{equation}

\subsubsection{Feature Final Preparation} \label{subsubsec:data_prep_feature_refinement}
First, the charging demand $\mathbf{d}_{t,i}$ is normalized, where missing values are initially imputed with zero. Using the mean $\mu_d$ and standard deviation $\sigma_d$ calculated from the temporal window, z-score standardization is applied, with the normalized demand written as
\begin{equation}
\label{eq:norm_demand}
    \mathbf{d}_{t,i}^\mathrm{no} = (\mathbf{d}_{t,i} - \mu_d) / \sigma_d.
\end{equation}

To ensure that the imputed zeros do not carry statistical meaning after normalization, the missing data mask $\mathbf{m}_{t,i}$ is used to reset the values at these positions back to zero. The masked normalized demand is denoted as $\mathbf{d}_{t,i}^\mathrm{ma,no}$.

Second, the calendar features $\mathbf{c}_t$ are encoded using sine and cosine transformations to preserve their cyclical nature, with outcomes denoted as $\mathbf{c}_{t}^\mathrm{sin}$ and $\mathbf{c}_{t}^\mathrm{cos}$, respectively, which are then concatenated to form the final calendar features, defined as $\tilde{\mathbf{c}}_t = [ \mathbf{c}_{t}^\mathrm{sin} || \mathbf{c}_{t}^\mathrm{cos} ] \in \mathbb{R}^6 $, where ``$[\cdot || \cdot]$'' represents the concatenation operation.

Finally, for station $i$ and day $t$, the complete set of features prepared for our PRAIM consists of the LLM embedding $\boldsymbol{\omega}_{t,i}$, the normalized, masked demand $\mathbf{d}_{t,i}^\mathrm{ma,no}$, and the cyclically encoded calendar features $\tilde{\mathbf{c}}_t$. It is noted that the prepared features after imputation can also be used for downstream forecasting task.

\section{PRAIM Imputation Framework} \label{sec:model_formulation}

This section details the architecture of our developed PRAIM--a conditional variational sequence-to-sequence imputation framework. This framework aims to learn a probabilistic mapping, denoted as $\mathbb{P} ( \hat{\mathbf{d}}_{t,i}^\mathrm{no} | \boldsymbol{\mathbf{d}}_{t,i}^\mathrm{ma,no}, \boldsymbol{\omega}_{t,i}, \tilde{\mathbf{c}}_t, v_i )$, from a set of conditioning variables (i.e., $\boldsymbol{\mathbf{d}}_{t,i}^\mathrm{ma,no}, \boldsymbol{\omega}_{t,i}, \tilde{\mathbf{c}}_t, v_i$) to a reconstructed sequence of charging demands, denoted as $\hat{\mathbf{d}}_{t,i}^\mathrm{no} \in \mathbb{R}^L$. The learned mapping is further characterized as a Gaussian distribution, defined as $\mathcal{N} ( \hat{\mathbf{d}}_{t,i}^\mathrm{no} | \hat{\boldsymbol{\mu}}_{t,i}, \hat{\boldsymbol{\sigma}}_{t,i}^2 )$, where $\mathcal{N}(\cdot|\cdot,\cdot)$ represents the Gaussian distribution, $\hat{\boldsymbol{\mu}}_{t,i}, \hat{\boldsymbol{\sigma}}_{t,i}^2\in\mathbb{R}^L$ are the learned mean and variance for the reconstructed charging demands, respectively.

To accurately learn the mapping, we move beyond simply concatenating all available information and feeding it into a single neural network. The critical drawback of existing concatenation-based approaches is that they simply treat all features as a flat vector, making it difficult to disentangle the hierarchical relationships between the high-level context (e.g., calendar information and PoIs) and the low-level time-series charging demand. Consequently, it could be challenging to adapt the resulted imputation strategy to the specific context of each data point. To strategically leverage contextual information, we design a more structured, multi-stage process that first focuses on enhancing the conditioning variables before using them to guide the imputation. Such enhancement is detailed in Section \ref{subsec:model_formulation_cond_enhancement}, including three key steps: 1) enriching the LLM-generated contextual embedding via RAG (to leverage statistical strengths from analogous scenarios and to make the derived embeddings more informative), 2) encoding the enriched context into a variational latent space (to capture and measure the ambiguity in the conditioning signal itself), and 3) using the resulting variational representation to generate adaptive parameters (to dynamically guide and shape the following charging demand reconstruction). The following Section \ref{subsec:model_formulation_trans} uses these parameters to modulate a transformer-based sequence decoder, outputting the learned Gaussian distribution.

\begin{figure}[!t]
    \centering
    \includegraphics[width=\linewidth]{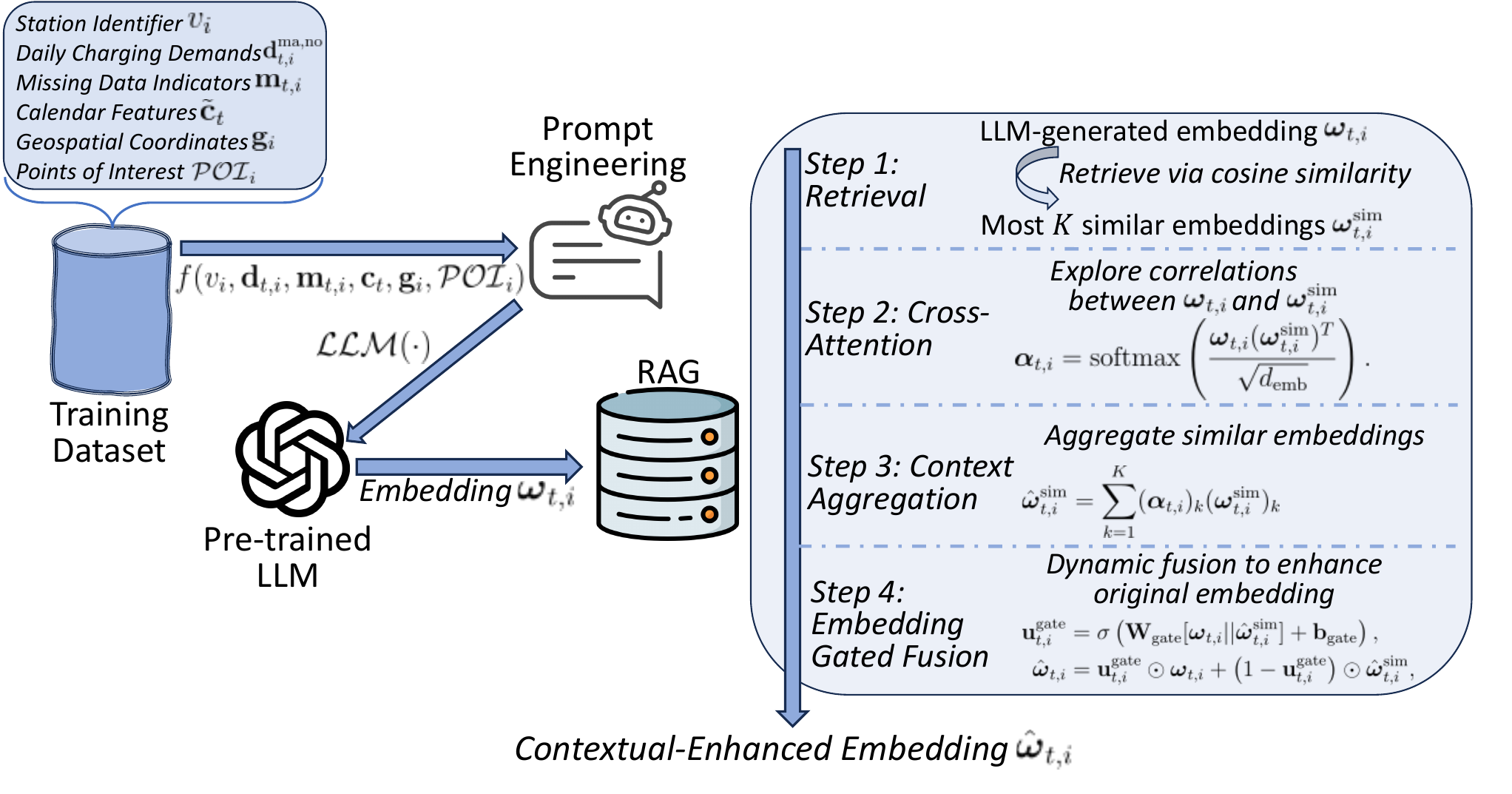}
    \caption{Illustration of RAG for embedding augmentation.}
    \label{fig:illustration_RAG}
\end{figure}

\subsection{Conditioning Variable Enhancement} \label{subsec:model_formulation_cond_enhancement}

\subsubsection{Retrieval-Augmented Generation} \label{subsubsec:model_formulation_cond_enhancement_rag}

We introduce RAG \cite{rag} to dynamically enhance LLM-generated embeddings. The motivation for using RAG is to mitigate data sparsity and improve generalization by providing relevant contextual evidence. For a given target station with a sparse or noisy history, its own embedding may not be sufficient for accurate imputation. To address this problem, we employ RAG as a non-parametric memory, which can retrieve embeddings from other stations or time periods that are semantically similar (e.g., sharing similar location types, calendar contexts, or demand patterns), even if they are not temporally or geographically adjacent. By fusing the original embedding with these relevant examples, we can derive a richer, more robust context with borrowed statistical strengths from analogous situations that RAG has recorded. Such refinement makes the imputation more informed and less reliant on potentially flawed local data.

Firstly, the embedding $\boldsymbol{\omega}_{t,i}$ serves as a query to retrieve the most similar $K$ embeddings from the RAG corpus. The retrieved embeddings are written as $\boldsymbol{\omega}_{t,i}^\mathrm{sim} = [ (\boldsymbol{\omega}_{t,i}^\mathrm{sim})_1 || (\boldsymbol{\omega}_{t,i}^\mathrm{sim})_2 || \cdots, || (\boldsymbol{\omega}_{t,i}^\mathrm{sim})_K ]$. 

Subsequently, the retrieved information is fused with the original query using a cross-attention mechanism, with corresponding attention weights $\boldsymbol{\alpha}_{t,i} \in \mathbb{R}^{K}$ calculated as
\begin{equation}
\label{eq:cross_att}
    \boldsymbol{\alpha}_{t,i} = \mathrm{softmax}\left( \frac{\boldsymbol{\omega}_{t,i} (\boldsymbol{\omega}_{t,i}^\mathrm{sim})^T}{\sqrt{d_\mathrm{emb}}} \right).
\end{equation}
The computed weights are used to produce a context vector in a weighted-sum manner, formulated as $\hat{\boldsymbol{\omega}}_{t,i}^\mathrm{sim} = \sum_{k=1}^{K} (\boldsymbol{\alpha}_{t,i})_k (\boldsymbol{\omega}_{t,i}^\mathrm{sim})_k$.

Further, a gated fusion mechanism combines the original embedding with the context vector to generate the refined embedding denoted as $\hat{\boldsymbol{\omega}}_{t,i}$. This mechanism is expressed as
\begin{align}
\label{eq:gated_fusion1}
    \mathbf{u}_{t,i}^\mathrm{gate} &= \sigma \left( \mathbf{W}_\mathrm{gate}[\boldsymbol{\omega}_{t,i} || \hat{\boldsymbol{\omega}}_{t,i}^\mathrm{sim}] + \mathbf{b}_\mathrm{gate} \right), \\
\label{eq:gated_fusion2}
    \hat{\boldsymbol{\omega}}_{t,i} &= \mathbf{u}_{t,i}^\mathrm{gate} \odot \boldsymbol{\omega}_{t,i} + \left( 1-\mathbf{u}_{t,i}^\mathrm{gate} \right) \odot \hat{\boldsymbol{\omega}}_{t,i}^\mathrm{sim},
\end{align}
where $\odot$ represents the Hadamard product operator, $\sigma(\cdot)$ is the sigmoid function, $\mathbf{W}_\mathrm{gate}$ and $\mathbf{b}_\mathrm{gate}$ are learnable parameters.

An illustration of how RAG operates is presented in Fig. \ref{fig:illustration_RAG}.

\begin{remark}[Adapativity to Concept Drift]
    Benefited from the non-parametric characteristic at the retrieval step, RAG thus can adapt to evolving behavior through lightweight corpus maintenance rather than structural changes. Adding new charging stations or newly observed days simply amounts to generating their embeddings via LLM and appending them to the RAG corpus. Meanwhile, the approximate-nearest-neighbor index can be updated incrementally. In addition, to address policy shifts or seasonality not present in the initial training data, the above RAG maintenance can be coupled with time-aware retrieval policies and the gating fusion mechanism already present in the current RAG.
\end{remark}

\subsubsection{Variational Latent Space} \label{subsubsec:model_formulation_cond_enhancement_vae}
The refined embedding is further projected to a variational latent space.The rationale behind is that the latent space acknowledges that a given context (e.g., a specific station on a Tuesday) can lead to a range of plausible demand patterns, not just one fixed outcome. Instead of concatenating the refined embedding directly with the input demands, the latent space can reflect how much variability we historically observe for ``days like this'' at ``stations like this'', given the station identity and its neighborhood type (from PoIs), the recent demand trajectory and which days are missing, and the calendar. When contexts are informative and consistent (e.g., a workplace PoI profile on a regular Tuesday with a typical last-week usage), the latent posterior concentrates and presents lower uncertainty. When contexts are ambiguous or conflicting (e.g., mixed PoI signals or long missing stretches), the latent posterior broadens and subsequently exhibits higher uncertainty. By encoding this variability, the designed variational space can represent ambiguity and is less prone to over-fitting. 

The variational space is built on a neural network $q(\mathbf{z}_{t,i} | \hat{\boldsymbol{\omega}}_{t,i})$ to learn the posterior distribution of the latent variable $\mathbf{z}_{t,i} \in \mathbb{R}^{d_\mathrm{lat}}$ given the embedding, where $d_\mathrm{lat}$ denotes the latent space dimension. The network can be parameterized as a diagonal Gaussian distribution $\mathcal{N} ( \mathbf{z}_{t,i} | \boldsymbol{\mu}_{t,i}^\mathbf{z}, \mathrm{diag}((\boldsymbol{\sigma}_{t,i}^\mathbf{z})^2) )$, where the mean $\boldsymbol{\mu}_{t,i}^\mathbf{z}$ and the variance $(\boldsymbol{\sigma}_{t,i}^\mathbf{z})^2$ are produced by two separate trainable linear transformations.

The approximated posterior is a multivariate normal distribution with a diagonal covariance matrix, allowing each latent dimension to be treated independently as
\begin{equation}
    \begin{aligned}
        q (\mathbf{z}_{t,i} | \hat{\boldsymbol{\omega}}_{t,i}) = \prod_{j=1}^{d_{\mathrm{lat}}} \mathcal{N} \left( (\mathbf{z}_{t,i})_j | (\boldsymbol{\mu}_{t,i}^\mathbf{z})_j, (\boldsymbol{\sigma}_{t,i}^\mathbf{z})^2_j \right).
    \end{aligned}
\end{equation}

To enable gradient back-propagation, we use the reparameterization trick formulated below to sample the latent variable,
\begin{equation}
\label{eq:reparameterization}
    \mathbf{z}_{t,i} = \boldsymbol{\mu}_{t,i}^\mathbf{z} + \boldsymbol{\sigma}_{t,i}^\mathbf{z} \odot \boldsymbol{\epsilon}.
\end{equation}

A sample $\mathbf{z}_{t,i}$ is drawn by first sampling a standard Gaussian random vector $\boldsymbol{\epsilon} \sim \mathcal{N}(\mathbf{0}, \mathbf{I})$ and then computing $\mathbf{z}_{t,i}$ deterministically based on the generated mean and variance.

\begin{remark}[Data Integrity in RAG]
    Benefiting from the LLM-generated high-dimensional embeddings and designed variational space, the proposed RAG does not lead to information leakage of time-aligned ground-truth charging demand of other stations or specific station-identifying records. Furthermore, to support more sensitive private data imputation, the RAG, without ever being exposed to the specific charging records, can enable privacy-preserving retrieval on network-wide charging patterns by incorporating techniques, such as differential privacy.
\end{remark}

\subsubsection{Feature-wise Linear Modulation} \label{subsec:model_formulation_film}

As mentioned at the beginning of this section, we first enhance conditioning variables before incorporating them to guide the imputation. The previous Sections \ref{subsubsec:model_formulation_cond_enhancement_rag} and \ref{subsubsec:model_formulation_cond_enhancement_vae} have detailed the refinement on the LLM-generated embedding. To complete the conditioning mechanism, we here introduce feature-wise linear modulation (FiLM) \cite{FiLM} to first fuse all available contextual information (including $\mathbf{z}_{t,i}$, $\hat{\mathbf{c}}_{t}$, and $v_i$) and then generate parameters to control imputation.

Firstly, the station identifier $v_i$ and calendar features $\hat{\mathbf{c}}_t$ are projected into high-dimensional spaces. For a given station identifier $v_i$, the corresponding embedding $\mathbf{h}_{\mathrm{stat}} \in \mathbb{R}^{d_{\mathrm{stat}}}$ is derived from a learnable embedding matrix $\mathbf{W}_{\mathrm{stat}} \in \mathbb{R}^{N_\mathrm{stat} \times d_{\mathrm{stat}}}$, where $N_\mathrm{stat}$ is the total number of charging stations and $d_\mathrm{stat}$ is the embedding dimension for station identifiers. Similarly, the calendar features $\tilde{\mathbf{c}}_t$ are projected via a neural linear layer, resulting in its embedding $\mathbf{h}_{t}^\mathrm{cal} \in \mathbb{R}^{d_\mathrm{cal}}$, where $d_\mathrm{cal}$ is the embedding dimension of calendar features.

These embeddings are concatenated with the latent variable $\mathbf{z}_{t,i}$ to form a unified conditioning vector, expressed as
\begin{equation}
    \mathbf{h}_{t,i}^{\mathrm{cond}} = [\mathbf{z}_{t,i} || \mathbf{h}_{\mathrm{stat}} ||  \mathbf{h}_{t}^\mathrm{cal}] \in \mathbb{R}^{d_{\mathrm{lat}} + d_{\mathrm{stat}} + d_{\mathrm{cal}}},
\end{equation}
which is then processed by a two-layer multi-layer perceptron (MLP) to produce two conditioning parameters $\boldsymbol{\gamma}_{t,i}, \boldsymbol{\beta}_{t,i} \in \mathbb{R}^{d_{\mathrm{FiLM}}}$, where $d_{\mathrm{FiLM}}$ is the dimension of the FiLM's output. The MLP can be broken down into the following structures.
\begin{enumerate}
    \item First linear projection: $\mathbf{h}_{t,i}^{\mathrm{hidd}} = \mathbf{W}_{\mathrm{cond1}} \mathbf{h}_{t,i}^{\mathrm{cond}} + \mathbf{b}_{\mathrm{cond1}}$, where $\mathbf{W}_{\mathrm{cond1}}\in\mathbb{R}^{ 2d_\mathrm{FiLM} \times (d_{\mathrm{lat}} + d_{\mathrm{stat}} + d_{\mathrm{cal}}) }$ and $\mathbf{b}_{\mathrm{cond1}} \in \mathbb{R}^{2d_\mathrm{FiLM}}$ are learnable parameters.
    
    \item Nonlinear activation: $\mathbf{h}_{t,i}^\mathrm{hidd} \leftarrow \mathbf{h}_{t,i}^\mathrm{hidd} \cdot \sigma(\mathbf{h}_{t,i}^\mathrm{hidd})$.
    
    \item Second linear projection: $\mathbf{h}_{t,i}^{\mathrm{out}} = \mathbf{W}_{\mathrm{cond2}} \mathbf{h}_{t,i}^{\mathrm{hidd}} + \mathbf{b}_{\mathrm{cond2}}$, where $\mathbf{W}_{\mathrm{cond2}} \in \mathbb{R}^{2d_\mathrm{FiLM} \times  2d_\mathrm{FiLM}}$ and $\mathbf{b}_{\mathrm{cond2}}\in \mathbb{R}^{2d_\mathrm{FiLM}}$.

    \item Splitting the output: The resulting vector $\mathbf{h}_{t,i}^{\mathrm{out}}$ is split into two halves for $\boldsymbol{\gamma}_{t,i}$ and $\boldsymbol{\beta}_{t,i}$.
\end{enumerate}

\subsection{Transformer-based Sequence Decoder} \label{subsec:model_formulation_trans}

The above conditioning parameters are further integrated into the following transformer-based decoder to derive the Gaussian distribution of the reconstructed charging demands. 

The aim of employing the decoder is to model temporal patterns within the input sequence of charging demands conditioned on contextual information. Understanding temporal context (e.g., a value missing on a Friday may depend on the value of the previous Friday) is essential for accurate imputation. This need can be effectively met by the transformer's self-attention mechanism that extracts temporal correlations between every pair of time steps in the sequence \cite{wang2025survey}. Thus, empowered by the transformer, our decoder can effectively look at all available charging demands in the temporal window to infer the most probable values for the missing ones.

The input daily charging demand $\mathbf{d}_{t,i}^\mathrm{ma,no}$ is first embedded by learnable parameters $\mathbf{W}_{\mathrm{emb}}$, $\mathbf{b}_{\mathrm{emb}}$, and then modulated by the FiLM parameters (i.e., $\boldsymbol{\gamma}_{t,i}$ and $\boldsymbol{\beta}_{t,i}$) as follows.
\begin{equation}
    \mathbf{d}_{t,i}^\mathrm{FiLM} = \boldsymbol{\gamma}_{t,i} \odot (\mathbf{W}_{\mathrm{emb}} \mathbf{d}_{t,i}^\mathrm{ma,no} + \mathbf{b}_{\mathrm{emb}}) + \boldsymbol{\beta}_{t,i}.
\end{equation}
Rather than concatenating the context with demands as the decoder's input, the above dynamic modulation enables a feature-wise affine transformation on the demands using contextual FiLM-generated parameters to reshape the scale ($\boldsymbol{\gamma}_{t,i}$) and shift ($\boldsymbol{\beta}_{t,i}$). For example, for a station identified as a workplace, the FiLM may learn to generate parameters that amplify weekday charging. This adaptive conditioning primes the input sequence, making the decoder perform more targeted and effective imputation.

\begin{figure*}[!t]
    \centering
    \includegraphics[width=.75\linewidth]{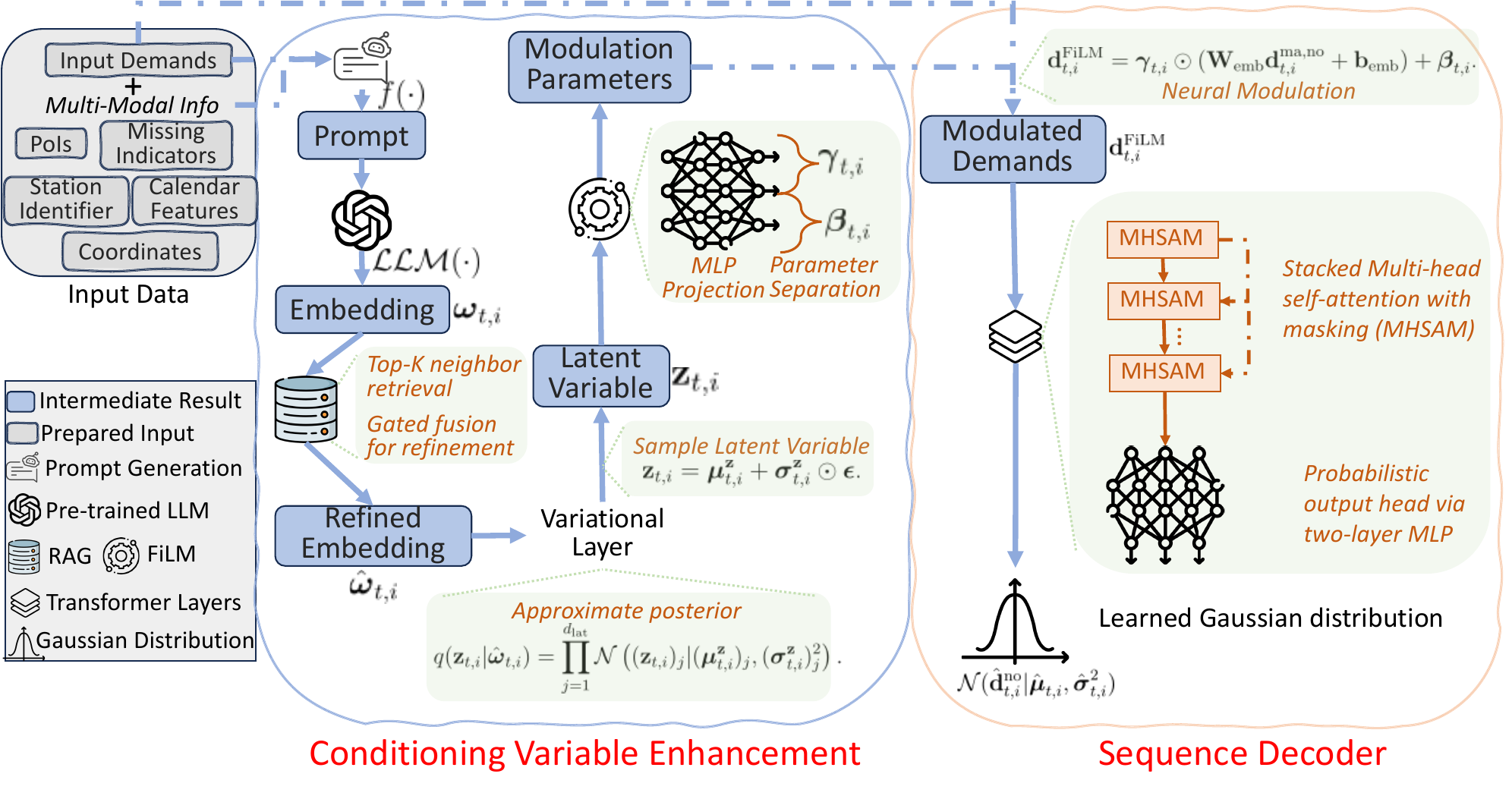}
    \caption{The workflow of PRAIM. Inputs per station $i$ and day $t$: $v_i$ (station ID), $\mathbf{m}_{t,i}$ (missing data indicators), $\tilde{\mathbf{c}}_t$ (calendar), $\mathbf{g}_i$ (coordinates), $\mathcal{POI}_{t,i}$, and $\mathbf{d}_{t,i}^\mathrm{ma, no}$ (historical demands). The LLM generates an embedding $\boldsymbol{\omega}_{t,i}$ based on the structured prompt, further refined by RAG as $\hat{\boldsymbol{\omega}}_{t,i}$. The following variational layer samples the latent variable $\mathbf{z}_{t,i}$ based on the posterior $q(\mathbf{z}_{t,i} | \hat{\boldsymbol{\omega}}_{t,i})$, which is fed into FiLM to produce parameters, $\boldsymbol{\gamma}_{t,i}$ and $\boldsymbol{\beta}_{t,i}$, to modulate the input demands into $\boldsymbol{\gamma}_{t,i}$ and $\boldsymbol{\beta}_{t,i}$. The final sequence decoder maps the modulated demands into the Gaussian distribution of reconstructed charging demands: $\mathcal{N}( \hat{\mathbf{d}}_{t,i}^\mathrm{no} | \hat{\boldsymbol{\mu}}_{t,i}, \hat{\boldsymbol{\sigma}}_{t,i}^2 )$.}
    \label{fig:model_workflow}
\end{figure*}

In addition to input modulation, our transformer-based decoder consists of a series of transformer layers for temporal modeling. Let $N_\mathrm{tran}$ denote the number of stacked layers. The core part of each layer is the \textit{multi-head self-attention with masking} (MHSAM), detailed as follows. For the $j$-th layer (where $j\in \{1,\cdots, N_\mathrm{tran}\}$), its input $\mathbf{d}_{t,i}^{(j-1)}$ is linearly projected into query, key, and value matrices for each attention head, written as $\mathbf{Q}_{t,i}^{(j, k)}$, $\mathbf{K}_{t,i}^{(j, k)}$, and $\mathbf{V}_{t,i}^{(j, k)}$, respectively, where $k$ represents the index of attention heads. Note that the input for the first layer is the modulated result $\mathbf{d}_{t,i}^\mathrm{FiLM}$ further processed by positional encoding \cite{bansal2021_Transformer}. The output of each head is calculated using scaled dot-product attention formulated as
\begin{equation}
    \boldsymbol{\kappa}_{t,i}^{(j, k)} = \mathrm{softmax} \left(\frac{\mathbf{Q}_{t,i}^{(j, k)} (\mathbf{K}_{t,i}^{(j, k)})^T}{\sqrt{d_\mathrm{FiLM}}} + \mathbf{M}_{t,i} \right)\mathbf{V}_{t,i}^{(j, k)} , 
\end{equation}
where the matrix $\mathbf{M}_{t,i} \in \mathbb{R}^{L\times L}$ is built on the missing data vector $\mathbf{m}_{t,i}$, preventing the decoder from attending to positions with missing data. This is achieved by assigning values of negative infinity to these positions, thereby forcing the softmax output probabilities for any masked position to become zero.

The outputs of all attention heads are concatenated and further fed into a two-layer MLP \cite{bansal2021_Transformer} for the output of the $j$-th layer. The output of the last transformer layer is passed through two separate linear heads to produce the mean and variance of the learned Gaussian distribution $\mathcal{N} ( \hat{\mathbf{d}}_{t,i}^\mathrm{no} | \hat{\boldsymbol{\mu}}_{t,i}, \hat{\boldsymbol{\sigma}}_{t,i}^2 )$, from which the reconstructed demand can be drawn. The key advantage of a probabilistic output over a single point estimate is the ability to quantify uncertainties of charging demand. By predicting both the mean and variance, our PRAIM delivers not only its best guess for a missing value but also its confidence in that guess. A low variance indicates high confidence, while a high variance signals that the reconstructed demand for that time step might be inherently noisy or unpredictable, which may include potential outages or communication issues. In addition, the retrieval-augmented context is able to reduce the risk of spuriously imputing non-zero values when zero is plausible. Specifically, PRAIM is conditioned on station-specific and cross-station context via RAG, which leverages strength from similar days/stations; when similar contexts indicate low or zero activity, the output mean tends toward low values, and uncertainty can increase when evidence is weak.

In summary, PRAIM leverages a rich contextual embedding refined by RAG. The enhanced embedding is encoded into a variational latent space to approximate the posterior of the latent variable and capture the embedding's uncertainty. The FiLM then conditions a transformer-based sequence decoder on this latent representation, along with station-specific information, enabling probabilistic imputation of the missing demand values. The diagram of PRAIM is illustrated in Fig. \ref{fig:model_workflow}.

\begin{remark}[LLM Hallucination]
    Unlike common approach directly using frozen/fine-tuned LLMs to complete certain tasks, the LLM in our PRAIM is not used to generate imputed values. The assistant role of LLM in producing contextual embeddings and the supervised training paradigm can reduce the impact of LLM hallucination on our model. Also, RAG has been demonstrated as an effective technique to mitigate such hallucination. In our PRAIM, RAG is used to refine and further contextualize the LLM-generated embeddings, leveraging statistical strength from real, observed patterns, thereby de-risking the hallucination. Further, our PRAIM quantifies uncertainties at two levels, i.e., the variational latent space and the probabilistic output, which bracket the LLM's role, making any residual hallucination measurable and controllable.
\end{remark}

\section{PRAIM Training and Evaluation} \label{sec:model_train_eval}

We outline training and evaluation procedures of PRAIM in Sections \ref{subsec:model_train_eval_train} and \ref{subsec:model_train_eval_eval}, respectively. 

The prepared EV charging dataset is denoted as $\mathcal{S}$, where each data point $\mathbf{s}_{t,i}$ is specified by the station index $i$ and time step $t$. To establish ground truth for imputation performance evaluation, we only include data points with a complete sequence of historical charging demands $\mathbf{d}_{t,i}$. The curated dataset is then partitioned into a training set $\mathcal{S}_\mathrm{train}$ and a test set $\mathcal{S}_\mathrm{test}$ based on a predetermined ratio. To test model imputation capability, we artificially mask the complete demand sequence. This process is controlled by a masking ratio, denoted as $\lambda$, which varies from $10\%$ to $90\%$ in increments of $10\%$. For each data point, we determine the number of time steps to mask by calculating $\lfloor \lambda\cdot L \rfloor $, where $\lfloor \cdot \rfloor$ is the floor function. Initially, the missing data indicator $\mathbf{m}_{t,i}$ is a vector of zeros. We then randomly select $\lfloor \lambda\cdot L \rfloor $ time steps within the temporal window and set their corresponding entries in $\mathbf{m}_{t,i}$ to one, thereby masking demands at those points. In Appendix B, we provide a detailed empirical analysis on how the random masking aligns with the realized missingness patterns compared to a distribution-matched masking.

With generated masks, each data point is structured as a combination of two sub-tuples. The first is the model input tuple $(v_i, \mathbf{d}_{t,i}, \mathbf{d}_{t,i}^{\mathrm{ma,no}}, \mathbf{m}_{t,i}, \tilde{\mathbf{c}}_t, \mathbf{g}_i, \mathcal{POI}_i )$, containing the features used to generate LLM embeddings and serving as inputs for the transformer-based decoder. The second is the ground truth tuple $(\mathbf{d}_{t,i}^\mathrm{no,mi}, \mathbf{d}_{t,i}^\mathrm{mi}, \mathcal{T}_{t,i}^\mathrm{mi})$, where $\mathbf{d}_{t,i}^\mathrm{no,mi}$ represents the normalized charging demands at the masked time steps, derived from the normalized demands $\mathbf{d}_{t,i}^\mathrm{no}$ by selecting values where mask $\mathbf{m}_{t,i}$ is one. $\mathbf{d}_{t,i}^\mathrm{mi}$ is the corresponding denormalized (i.e., original scale) version of the masked demands, recovered using the mean $\mu_d$ and standard deviation $\sigma_d$ from Eq. \eqref{eq:norm_demand}. $\mathcal{T}_{t,i}^\mathrm{mi}$ is the set of time steps at which demands are masked. In addition, we denote the learned Gaussian distribution at these masked time steps as $\mathcal{N}(\hat{\boldsymbol{\mu}}_{t,i}^\mathrm{mi}, (\hat{\boldsymbol{\sigma}}_{t,i}^\mathrm{mi})^2)$.

\subsection{Model Training}  \label{subsec:model_train_eval_train}

PRAIM is trained by optimizing a loss function known as the negative evidence lower bound (ELBO) \cite{xiao2023diffusion,tashiro2021diffusion} composed of two competing yet complementary parts: a reconstruction loss and a Kullback-Leibler (KL) divergence loss. The former pushes the model to generate outputs that are as close as possible to the ground truth. However, this stand-alone loss makes the model simply memorize the training examples, leading to over-fitting. To counteract this problem, the KL divergence loss serves as a regularizer, restricting the learned latent representation to be simple and well-organized. This loss prevents the model from creating an overly complex or sparse latent space, encouraging it to learn more general patterns. Hence, training with the ELBO is a balancing act, which simultaneously motivates the model to be accurate in its reconstruction while learning generalized internal representations. Such a dual-objective drives the model to perform well on both seen and unseen data.

\subsubsection{Reconstruction Loss}

Written as $\mathcal{L}_{\mathrm{rec}}(\cdot)$, it calculates the negative log-likelihood of the normalized ground truth (at masked time steps) $\mathbf{d}_{t,i}^\mathrm{no,mi}$ given the learned distribution $\mathcal{N}(\hat{\boldsymbol{\mu}}_{t,i}^\mathrm{mi}, (\hat{\boldsymbol{\sigma}}_{t,i}^\mathrm{mi})^2)$, whose probability density function given a time step $t'\in \mathcal{T}_{t}^\mathrm{mi}$ is written as
\begin{equation}\label{eq_density}
    \begin{aligned}
        &\mathbb{P}( (\mathbf{d}_{t,i}^\mathrm{no,mi})_{t'} | (\hat{\boldsymbol{\mu}}_{t,i}^\mathrm{mi})_{t'}, (\hat{\boldsymbol{\sigma}}_{t,i}^\mathrm{mi})^2_{t'}) \\
        &= \frac{1}{\sqrt{2\pi(\hat{\boldsymbol{\sigma}}_{t,i}^\mathrm{mi})^2_{t'}}} \exp\left(-\frac{((\mathbf{d}_{t,i}^\mathrm{no,mi})_{t'}-(\hat{\boldsymbol{\mu}}_{t,i}^\mathrm{mi})_{t'})^2}{2(\hat{\boldsymbol{\sigma}}_{t,i}^\mathrm{mi})^2_{t'}}\right).
    \end{aligned}
\end{equation}

Applying negative natural logarithm to the above density function Eq.~\eqref{eq_density} gives the reconstruction loss of one single time step, and averaging over $\mathcal{T}_{t}^\mathrm{mi}$ gives the loss per data point as
\begin{equation}
    \begin{aligned}
        \mathcal{L}_{\mathrm{rec}} = \frac{1}{|\mathcal{T}_{t}^\mathrm{mi}|} \sum_{t'\in \mathcal{T}_{t}^\mathrm{mi}} \left[ \frac{((\mathbf{d}_{t,i}^\mathrm{no,mi})_{t'}- (\hat{\boldsymbol{\mu}}_{t,i}^\mathrm{mi})_{t'})^2}{2(\hat{\boldsymbol{\sigma}}_{t,i}^\mathrm{mi})^2_{t'}} \right. \\
        \left.  + \frac{1}{2}\log((\hat{\boldsymbol{\sigma}}_{t,i}^\mathrm{mi})^2_{t'}) + \frac{1}{2}\log(2\pi) \right].
    \end{aligned}
\end{equation}

\subsubsection{KL Divergence Loss} 

Denoted as $\mathcal{L}_{\mathrm{KL}}(\cdot)$, it regularizes the latent space by penalizing the divergence of the approximated posterior $q(\mathbf{z}_{t,i}| \hat{\boldsymbol{\omega}}_{t,i} )$, i.e., $\mathcal{N} ( \boldsymbol{\mu}_{t,i}^\mathbf{z}, \mathrm{diag}((\boldsymbol{\sigma}_{t,i}^\mathbf{z})^2) )$, from the standard Gaussian prior $ \mathcal{N}(\mathbf{0}, \mathbf{I})$. The analytical form for two diagonal Gaussians yields the KL loss, formulated as
\begin{equation}
    \begin{aligned}
        \mathcal{L}_{\mathrm{KL}} = \frac{1}{2} \sum_{j=1}^{d_{\mathrm{lat}}} \left[ (\boldsymbol{\sigma}_{t,i}^\mathbf{z})^2_j + (\boldsymbol{\mu}_{t,i}^\mathbf{z})^2_j - \log((\boldsymbol{\sigma}_{t,i}^\mathbf{z})^2_j) - 1 \right].
    \end{aligned}
\end{equation}
The ELBO is thus written as $\mathcal{L}_{\mathrm{ELBO}} = \mathcal{L}_{\mathrm{rec}} + \theta \cdot \mathcal{L}_{\mathrm{KL}}$, where $\theta$ is a weighting hyperparameter. The gradient descent is used to leverage the loss for model parameter update.

\subsection{Model Evaluation}  \label{subsec:model_train_eval_eval}

We assess PRAIM's imputation accuracy using the mean absolute error (MAE). Moreover, to ensure a consistent and deterministic evaluation, we eliminate randomness from the model's workflow as follows. First, we bypass the stochastic element of the latent space by setting the latent variable $\mathbf{z}_{t,i}$ directly to its learned mean $\boldsymbol{\mu}_{t,i}^\mathbf{z}$. Similarly, the final output is determined not by sampling from the learned Gaussian distribution, but by directly taking its mean $\hat{\boldsymbol{\mu}}_{t,i}$. Finally, this deterministic output is denormalized with the mean $\mu_d$ and the standard deviation $\sigma_d$ calculated in Eq. \eqref{eq:norm_demand} to produce the reconstructed demand sequence $\hat{\mathbf{d}}_{t,i}$ for metric calculations.

\section{Experiments and Results} \label{sec:experiments}

\subsection{Experimental Settings} \label{subsec:exp_settings}

\subsubsection{Benchmarks}

We compare our PRAIM against three categories of established benchmark methods.
\begin{itemize}
    \item \textit{Statistical} \cite{wang2025survey}: mean, zero, the last-observed (LO) value, and interpolation (IP) imputation.

    \item \textit{ML}: KNN \cite{wang2025survey}, matrix factorization (MaF) \cite{wang2025survey}, Kalman filter (KF) \cite{Demirhan2018kalman}, MissForest (MF) \cite{Stekhoven2011missforest}.

    \item \textit{DL}: LSTM \cite{lee2020lstm}, temporal convolutional network (TCN) \cite{fahim2025lstmcnn}, transformer (TF) \cite{bansal2021_Transformer}, GAN \cite{miao2021GAN}, denoising AE (DAE) \cite{liguori2023autoencoder}.

    \item \textit{LLM}: LLM4RS \cite{R5C5_1}, LLM4HRS \cite{R5C5_2}, LLM4Code \cite{R5C5_3}.
\end{itemize}

\subsubsection{Model Configurations}
The geographic radius for querying PoIs is set to $2$ kilometers. We use the open-source AngIE as the pre-trained LLM to generate embeddings with length $d_\mathrm{emb}$ of $4096$ \cite{liangle2024}. The temporal window length of charging demand sequence $L$ is set to $7$. For RAG, the number of searched nearest neighbors, i.e., $K$, is set to $20$.

\textit{Model Architecture}: The latent space dimension $d_\mathrm{lat}$ is set to $256$. The embedding dimensions for both station identifiers ($d_\mathrm{stat}$) and calendar features ($d_\mathrm{cal}$) are set to $32$. The output dimension of the FiLM $d_\mathrm{FiLM}$ is $256$. The transformer-based decoder consists of $4$ layers, each with $8$ attention heads.

\textit{Implementation}: The ratio of splitting training and test sets is $80:20$. The weight for the KL divergence loss $\theta$ is set to $0.1$. We use the Adam optimizer to update neural weights, with an initial learning rate of $1\mathrm{e}^{-4}$ and a weight decay factor of $1\mathrm{e}^{-5}$. The training batch size is $64$. An early stopping mechanism is introduced, alongside the non-parametric RAG, MHSAM, and KL regularization, to mitigate model overfitting. PRAIM is implemented using PyTorch 1.12.0 and Python 3.10.13. Code is available at \url{https://github.com/StephLee12/PRAIM}.

\begin{remark}[Model Extensibility to Finer Temporal Resolution]
    PRAIM is designed to be resolution-agnostic. The neural components and training logic remain the same, with only light adaptations to inputs and missing data indicators. Specifically, temporal aggregation is transformed from daily to hourly or minute-level (e.g., 30-minute) bins. The calendar features can be augmented with time-of-day tokens, and the geospatial context remains unchanged. The retrieval query naturally expands to include time-of-day context. Thus, the nearest retrieved neighbors reflect both place and intra-day position.
\end{remark}

\begin{figure*}[!t]
\centering
    \subfloat[Palo Alto, US]{
    \includegraphics[width=.88\linewidth]{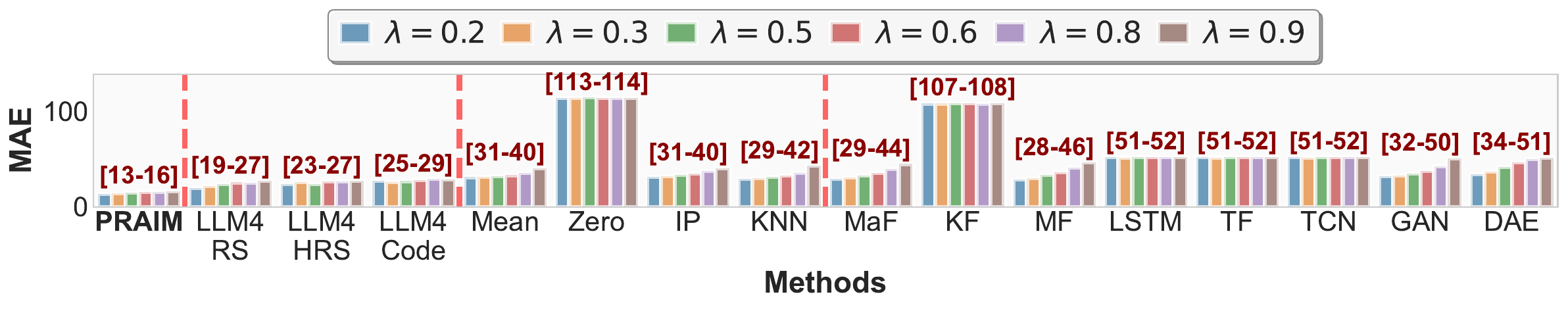}
    } \\ [-0.5mm]
    \subfloat[Boulder, US]{
    \includegraphics[width=.88\linewidth]{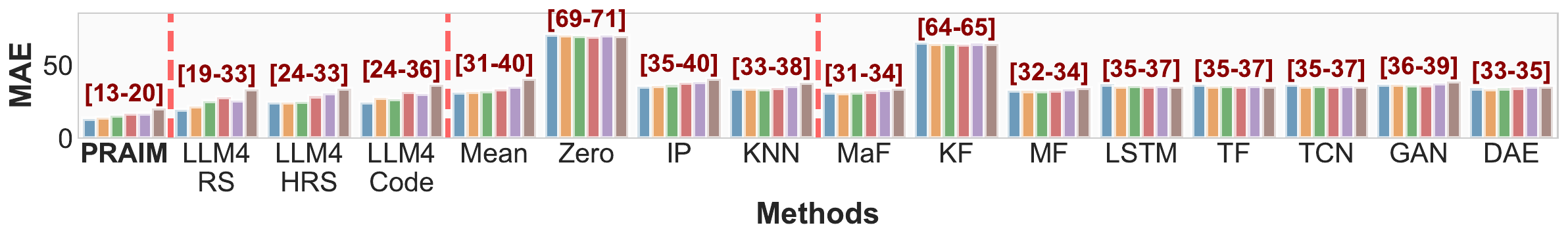}
    
    } \\ [-0.5mm]
    \subfloat[Dundee, UK]{
    \includegraphics[width=.88\linewidth]{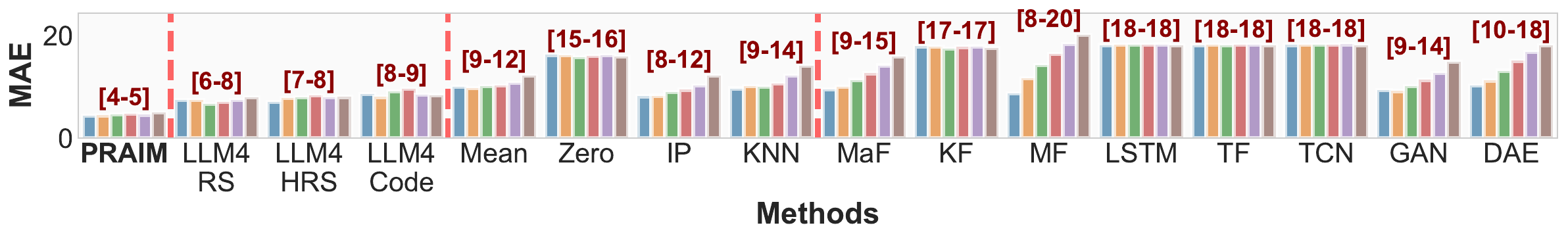}
    } \\ [-0.5mm]
    \subfloat[Perth, UK]{
    \includegraphics[width=.88\linewidth]{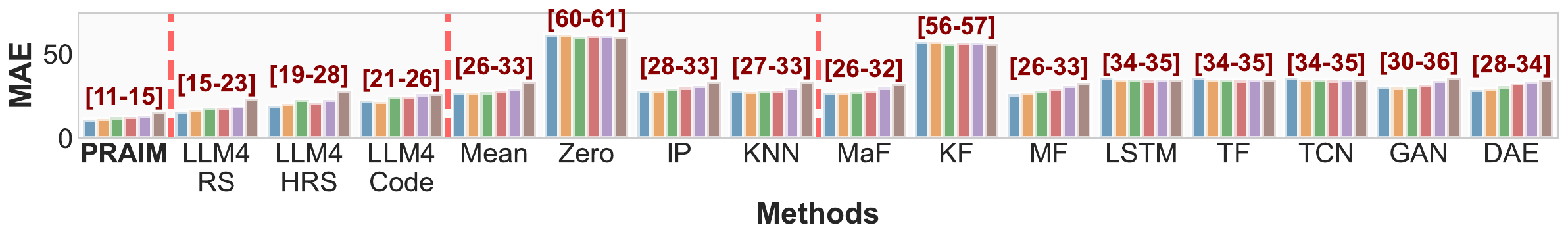}
    } 
    \caption{MAE comparison between PRAIM and benchmarks across used four datasets.}
    \label{fig:MAE_comparison}
\end{figure*}

\subsection{Empirical Results} \label{subsec:exp_results}

In this section, we structure our experimental results into three parts. First, Section \ref{subsubsec:exp_results_performance_comp} evaluates PRAIM's imputation accuracy against the previously mentioned benchmarks using artificially masked data. Section \ref{subsubsec:exp_results_imputation_actual_demand} then applies our model to impute the actual missing values in the datasets, providing a quantitative analysis of the imputed demand distributions. Finally, Section \ref{subsubsec:exp_results_downstream_forecast} investigates the practical utility of our method by evaluating the impact of the imputed data on the performance of a downstream demand forecasting task.

\subsubsection{Performance Comparison}  \label{subsubsec:exp_results_performance_comp}

We evaluate scenarios where the number of missing days within the $7$-day temporal window ranges from $1$ to $6$. The results are illustrated in Fig. \ref{fig:MAE_comparison}, where the clustered bars (for each method) from left to right present the MAE for an increasing number of missing days, i.e., from $1$ to $6$. Across all four datasets, PRAIM consistently achieves the lowest MAE compared to benchmark models and maintains this advantage across all levels of data absence, highlighting both the accuracy and robustness of our approach. These results can be attributed to the core design of our framework. Unlike benchmarks that rely heavily on the immediate temporal sequence, PRAIM leverages rich, multimodal context encoded by the LLM and dynamically augmented by the RAG. By retrieving relevant historical patterns from the charging network, PRAIM effectively compensates for the lack of local information, enabling highly accurate imputation even when the majority of the data points in a sequence are missing. These results validate the efficacy of our universal, context-aware imputation strategy.

Notably, LLM baselines consistently outperform classical statistical/ML/DL methods, suggesting the huge benefits of leveraging LLM in fusing heterogeneous station context (e.g., calendar features and PoIs) with missing demand history into a semantic-rich representation. Among the LLM family, LLM4RS is closest to PRAIM, followed by LLM4HRS and then LLM4Code. This ordering is intuitive and reasonable: once a pre-trained LLM is adapted to the target domain via fine-tuning, its embeddings align better with downstream reconstruction, narrowing the gap to PRAIM. Nevertheless, the remaining gap in favor of PRAIM demonstrates the value of our pipeline after the LLM/RAG stage.

The evaluation results from Fig. \ref{fig:MAE_comparison} suggest that the same approach (including our PRAIM and baselines) can produce different absolute and relative performance across datasets for reasons that are intrinsic to the data rather than the model, including 1) various missingness patterns, 2) different station's geospatial mix and urban morphology, 3) historical data coverage and data quality, and 4) policy and seasonality drifts by regions within the city and year.

\begin{table}[!t]
    \caption{Relative change in MAE under hyperparameter sensitivity analysis compared to the settings used in the current experiments.}
    
    \centering
    
    \resizebox{\columnwidth}{!}{
    \begin{tabular}{c|c|c|c|c|c}
    
    \multicolumn{2}{c|}{\diagbox[height=2\normalbaselineskip]{Parameter}{Dataset}} & Boulder & Dundee & Palo Alto & Perth  \\
    
    \hline 
    
    \multirow{2}{*}{$K$} & $10$ & $+10.2\%$ & $+9.6\%$ & $+10.8\%$ & $+10.1\%$ \\ 
    
    \cline{2-6}
    
    & $30$ & $-2.8\%$ & $-3.2\%$ & $-3.0\%$ & $-2.6\%$ \\
    
    \hline 
    
    $L$ & $14$ & $-1.8\%$ & $+2.7\%$ & $-3.5\%$ & $+1.9\%$ \\

    \hline 
    
    \multirow{2}{*}{$d_\mathrm{lat},d_\mathrm{FiLM}$} & $128$ & $+3.6\%$ & $+3.9\%$ & $+4.1\%$ & $+3.2\%$ \\ 
    
    \cline{2-6}
    
    & $512$ & $-1.2\%$ & $-0.8\%$ & $-1.5\%$ & $-0.9\%$ \\
    
    \hline 
    
    \multirow{2}{*}{$d_\mathrm{stat},d_\mathrm{cal}$} & $16$ & $+2.4\%$ & $+2.9\%$ & $+2.1\%$ & $+2.6\%$ \\ 
    
    \cline{2-6}
    
    & $64$ & $-0.6\%$ & $+0.5\%$ & $-1.0\%$ & $+0.3\%$ \\
    
    \hline 
    
    \multirow{2}{*}{$N_\mathrm{tran}$} & $2$ & $+10.7\%$ & $+9.8\%$ & $+11.2\%$ & $+10.3\%$ \\

    \cline{2-6}

    & $6$ & $-5.8\%$ & $-6.3\%$ & $-6.0\%$ & $-5.6\%$ \\

    \hline 
    
    \multirow{2}{*}{$\theta$} & $0.05$ & $-1.1\%$ & $+0.9\%$ & $-0.9\%$ & $+1.0\%$ \\

    \cline{2-6}

    & $0.2$ & $+1.4\%$ &  $-0.6\%$ & $+1.2\%$ & $-0.7\%$ \\

    \hline

    \end{tabular}
    }
    \label{tab:sensitivity_analysis}    
\end{table}

\textbf{Probabilistic Output Evaluation}: In addition to point-wise, deterministic imputation evaluation, i.e., MAE, we introduce metrics, including negative log-likelihood (NLL), continuous ranked probability score (CRPS), and interval coverage probability (ICP) to examine the effectiveness of our PRAIM. The detailed results are provided in Appendix C, which show that our PRAIM delivers consistently stronger probabilistic performance than the best-performing baseline (except the LLM baselines) for each dataset.

\textbf{Hyperparameter Sensitivity Analysis}: To further justify the chosen hyperparameter values, we have performed sensitivity analysis on the following parameters around the defaults: 1) $K\in \{10, 20, 30\}$, 2) $L \in \{7, 14\}$, 3) $d_\mathrm{lat}$, $d_\mathrm{FiLM}$ $\in \{128, 256, 512\}$, 4) $d_\mathrm{stat}$, $d_\mathrm{cal}$ $\in \{16, 32, 64\}$, 5) $N_\mathrm{tran} \in \{ 2, 4, 6 \}$, and 6) $\theta \in \{ 0.05, 0.1, 0.2 \}$. Compared to our current hyperparameter settings, the relative change in MAE is reported in Table \ref{tab:sensitivity_analysis}, where negative/positive values represent imputation performance improvement/degradation. The results show that PRAIM's accuracy is stable within practical bands, and our defaults sit near clear accuracy–efficiency knees. Specifically, the sensitivity analysis reveals that our default hyperparameter choices deliver consistent performance and lie near operationally sensible knees: moving to lighter settings incurs noticeable error (e.g., $K=10$, or $N_\mathrm{tran}=2$), while heavier settings offer only marginal gains at higher cost (e.g., $K=30$). The results further indicate that our method's performance gains are not artifact or narrow hyperparameter tuning but persist across reasonable ranges.

\textbf{Ablation Study}: The effectiveness of LLM-generated embeddings (grounded by RAG) is further demonstrated by replacing LLM (alongside RAG) with conventional encoders that still trail our PRAIM framework. Detailed results are provided in Appendix D.

Additionally, three advanced prompt engineering strategies \cite{R4C1_prompt_1} are introduced and compared to the current prompt (as shown in Fig. \ref{fig:example_prompt}). Detailed comparison results are provided in Appendix E, which confirms that better-engineered prompts do help but with marginal improvements (less than 10\% of decrease in relative MAE). The results indicate that 1) PRAIM's effectiveness is not tightly constrained by prompt engineering and 2) PRAIM's core, i.e., RAG grounding alongside variational conditioning with FiLM-modulated decoding, already provides robust performance.

Also, the impact of the filter threshold during the data preprocessing described in Section \ref{sec:dataset_overview} is analyzed in Appendix F, which empirically shows that setting the value of threshold of $35\%$ tend to filter out stations whose history is too unreliable while preserving enough cross-station heterogeneity to make RAG effective and training stable.

\subsubsection{Inference Time Cost Analysis}
The inference time of PRAIM per data point is less than $25$ milliseconds (ms), with LLM $13$–$15$ ms, RAG $5$–$7$ ms, and other parts (including the variational layer, FiLM, and decoder) $3$ ms. Lightweight LSTM/transformer baselines are around $1-4$ ms and reconstruction-based methods (e.g., GAN) range from $3$ to $10$ ms. Though our PRAIM is thus around $5-20$ times slower than baselines, such inference costs are fast enough for both rolling batch imputation and large-scale charging networks.

\subsubsection{Imputation on Actual Missing Demands} \label{subsubsec:exp_results_imputation_actual_demand}

Beyond evaluating performance on artificially masked data, it is crucial to assess how well an imputation method can reconstruct the underlying distribution of the actual missing records. An effective imputation method should generate values that are not only plausible but also statistically consistent with the observed data, thereby preserving the dataset's natural characteristics. To this end, we apply PRAIM and the benchmark models to fill the genuine missing entries in each dataset. We then compare the distribution of the original, non-missing data against the distribution of the imputed data \textit{for each charging station}.

This comparison is quantified using four statistical metrics. We measure the relative difference in the \textit{mean} and the \textit{coefficient of variation} to compare first and second-moment statistics. In addition, three distributional metrics are employed. The first is the \textit{Wasserstein distance}, which measures the minimum cost required to transform one distribution into the other. The second is the two-sample \textit{Kolmogorov-Smirnov (KS)} test, a non-parametric test that quantifies the maximum difference between the cumulative distribution functions of the two samples. The third is the \textit{Mann-Whitney U} test, another non-parametric test that assesses whether two independent samples are drawn from populations with the same distribution. For the KS and Mann-Whitney tests, we consider the distributions to be statistically similar if the test fails to reject the null hypothesis at a p-value threshold of $0.05$.

\begin{figure*}[!t]
\centering
    \subfloat[Palo Alto, US]{
    \includegraphics[width=.46\linewidth]{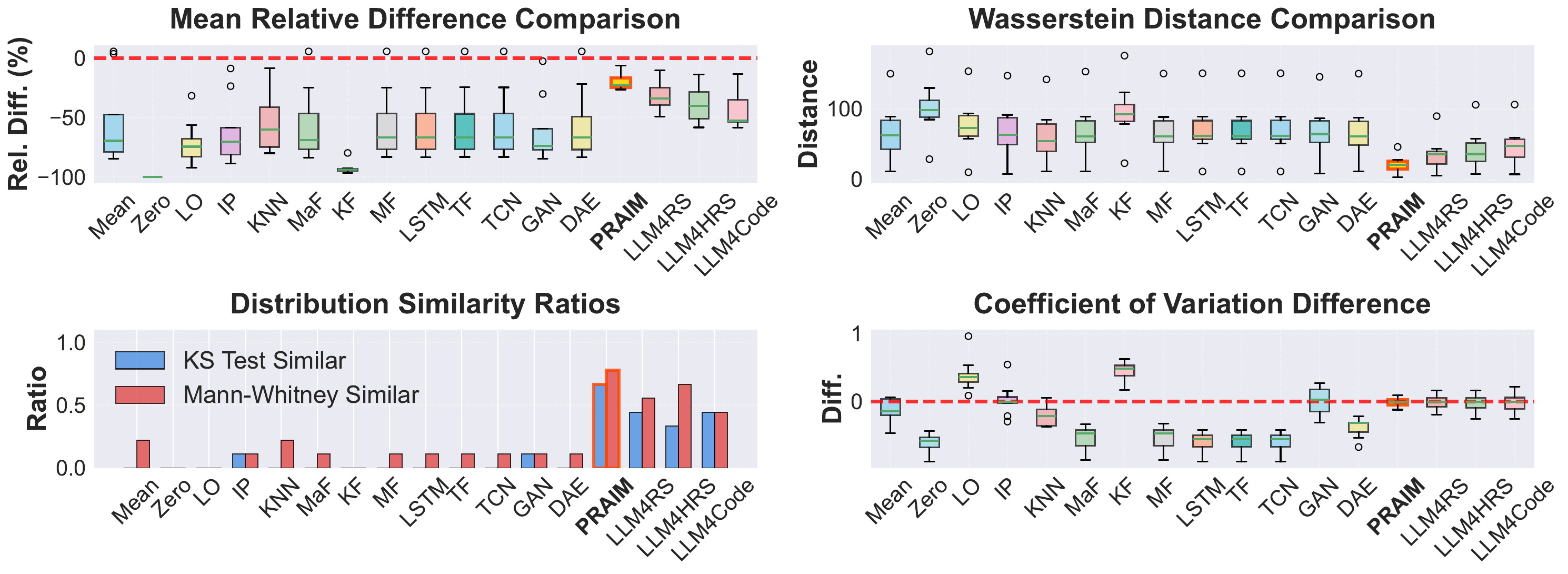}
    \label{fig:distribution_comparison_paloalto}
    }
    \subfloat[Boulder, US]{
    \includegraphics[width=.46\linewidth]{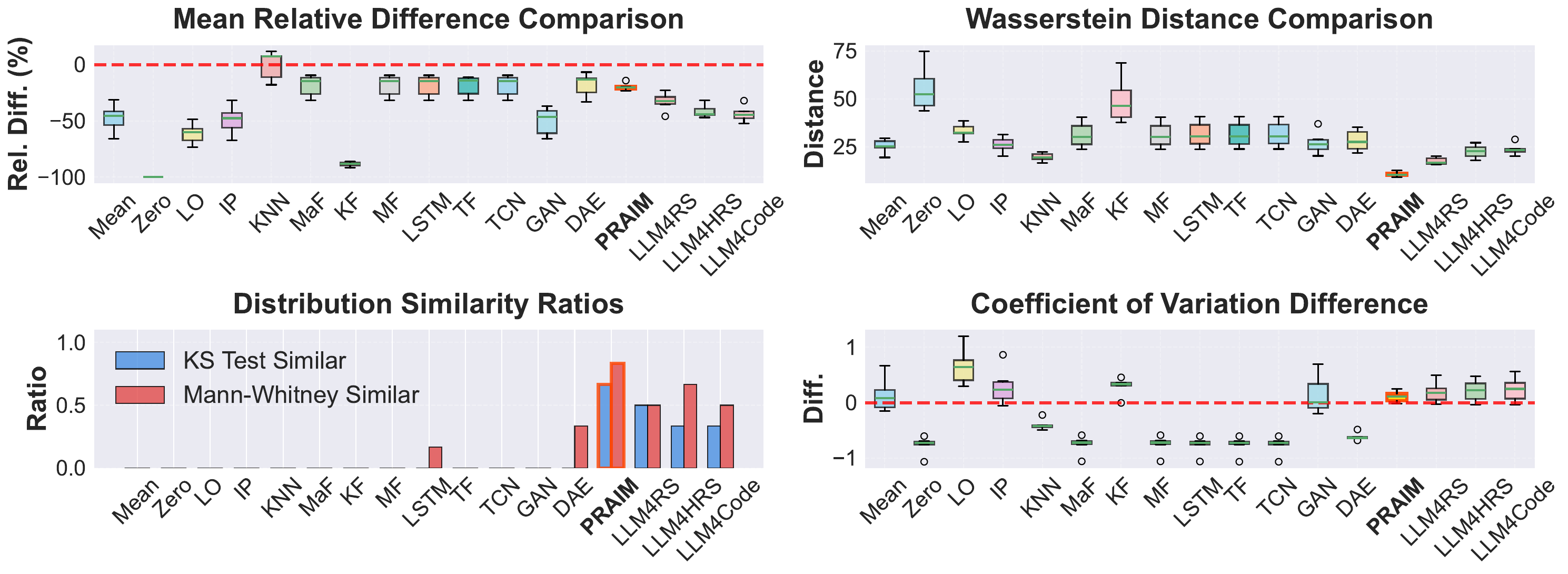}
    \label{fig:distribution_comparison_boulder}
    }\\ [-2mm]
    \subfloat[Dundee, UK]{
    \includegraphics[width=.46\linewidth]{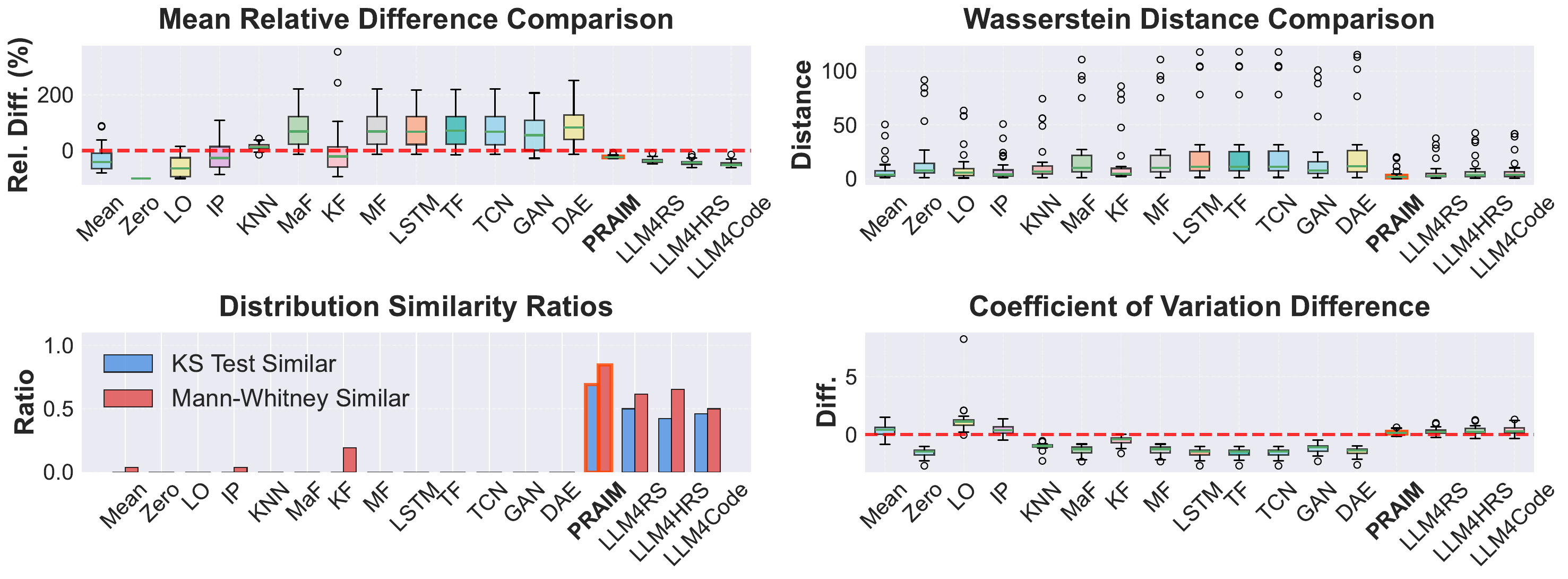}
    \label{fig:distribution_comparison_dundee}
    }
    \subfloat[Perth, UK]{
    \includegraphics[width=.46\linewidth]{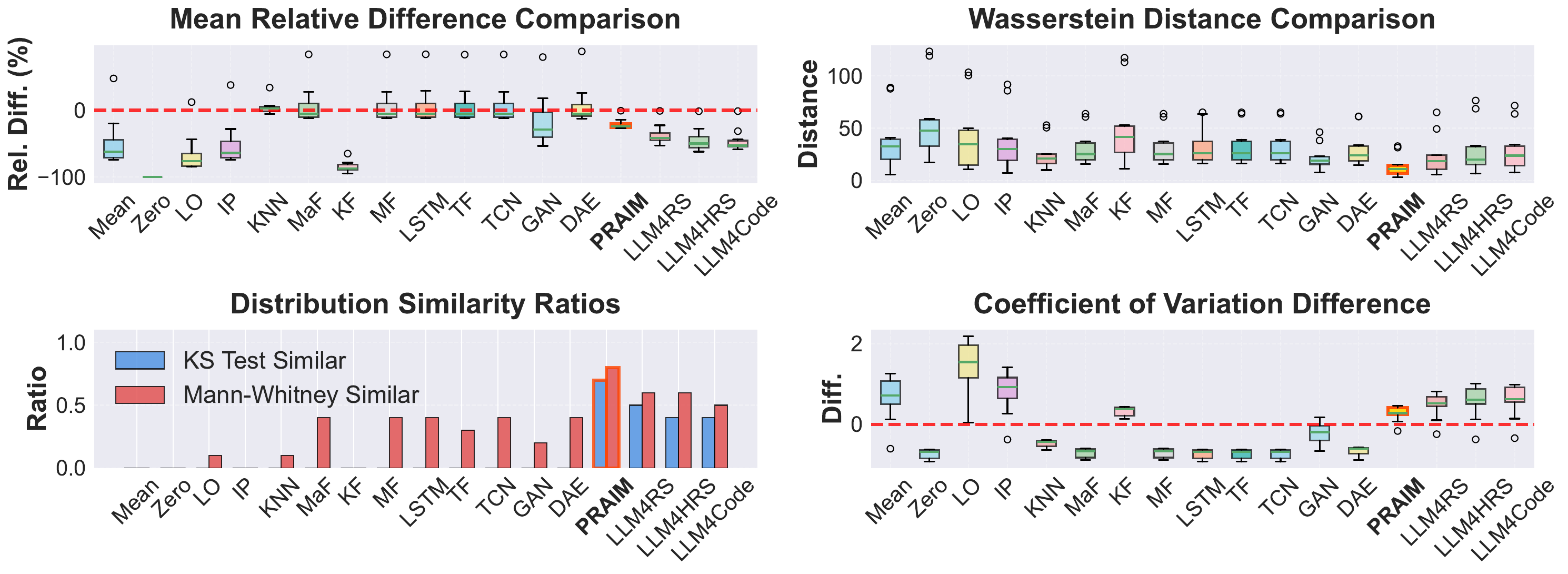}
    \label{fig:distribution_comparison_perth}
    }
    \caption{Distributional statistics comparison between imputed data distribution and original data distribution.}
    \label{fig:distribution_comparison}
\end{figure*}

The statistical results shown in Fig. \ref{fig:distribution_comparison} consistently favor PRAIM. Specifically, in the ``Mean Relative Difference'' and ``Coefficient of Variation Difference'' plots, PRAIM's boxplots are tightly centered around zero across all four datasets, indicating that the mean and relative variability of the data imputed by PRAIM are significantly similar to those of the original data. In contrast, other methods often introduce significant bias, either under or overestimating these fundamental statistics.

Moreover, the ``Wasserstein Distance Comparison'' plots show that PRAIM achieves the lowest distance values by a significant margin, highlighting that the overall shape of its imputed distribution is closest to the original. Furthermore, the ``Distribution Similarity Ratios'' plots reveal that for a high proportion of charging stations (over $75\%$ in most cases), the data imputed by PRAIM is statistically indistinguishable from the original data according to both the KS and Mann-Whitney U tests. No other method comes close to this level of distributional fidelity; for most benchmarks, the imputed data are statistically different from the original for nearly all stations.

\begin{figure}[!t]
    \centering
    \includegraphics[width=\linewidth]{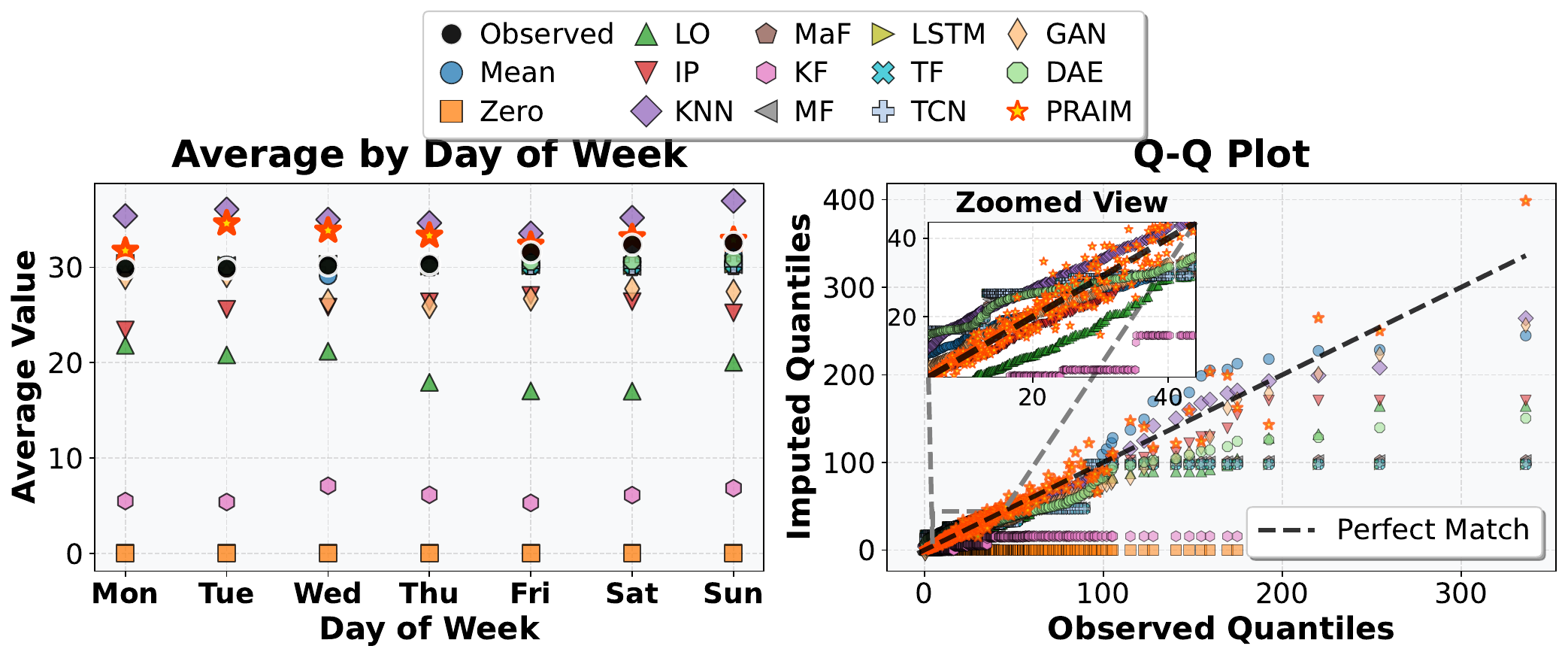}
    \caption{Day-of-week and quantile analysis of imputed data distribution.}
    \label{fig:dayofweek_and_quantile_analysis}
\end{figure}

Additionally, taking Palo Alto as an example, Fig. \ref{fig:dayofweek_and_quantile_analysis} provides a granular look at the imputed distributions. The ``Average by Day of Week'' plots show that the values imputed by PRAIM (yellow star) accurately capture the underlying weekly demand patterns, closely matching the observed averages (black circles) for each day of week. In contrast, most other advanced models fail to replicate the daily nuances with the same fidelity. Further, the quantile-quantile (Q-Q) plots reinforce this conclusion. The points generated by PRAIM align almost perfectly with the diagonal ``Perfect Match'' line, suggesting that the imputed data tends to share the same distribution as the observed data across all quantiles. These visual results provide strong intuitive support for the quantitative findings from our statistical tests.

In summary, these findings demonstrate that PRAIM outperforms not only in point-wise imputation accuracy but also in its ability to faithfully reconstruct the statistical properties of the true data distribution. This capability ensures that the completed dataset is a reliable representation of real-world charging behavior, making it suitable for downstream analysis.

\subsubsection{Impacts on Downstream Forecasting} \label{subsubsec:exp_results_downstream_forecast}

The final test of an imputation method is its ability to enhance the performance of downstream applications. We here evaluate the utility of the imputed data in EV charging demand forecasting task as a case study. The task is to predict the charging demand for the next three days using the previous seven days of data as input. We implement several state-of-the-art time-series forecasting models using the \texttt{NeuralForecast} library \cite{neuralforecast}.

\begin{figure}[!t]
    \centering
    \includegraphics[width=0.9\linewidth]{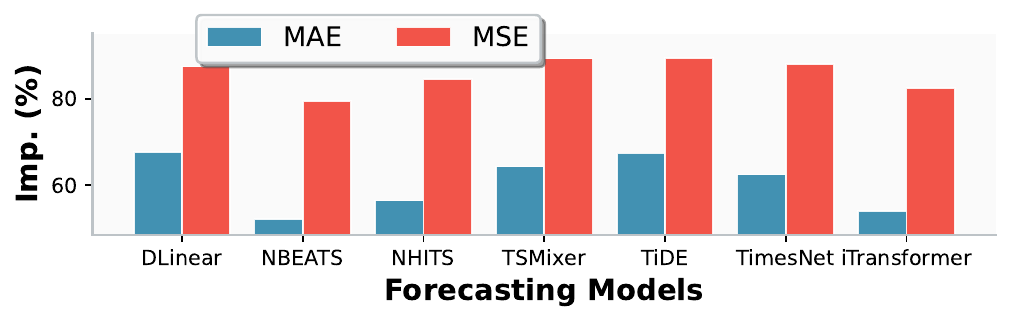}
    \caption{Forecasting performance using imputed charging datasets.}
    \label{fig:forecast_performance}
\end{figure}

We measure the performance improvement by calculating the percentage decrease in MAE and mean squared error (MSE) when the forecasting models are trained on the full, imputed dataset compared to being trained on only the original, incomplete data. Note that the two metrics are only calculated using predictions that have corresponding ground truth (i.e., not imputed data). The results of Palo Alto dataset are shown in Fig. \ref{fig:forecast_performance}, demonstrating the significant positive impact of our imputation method. Across all datasets and for every forecasting model tested, using the PRAIM-completed data leads to a notable reduction in forecasting errors. A crucial insight from the figure is that the percentage improvement in MSE is consistently much larger than the improvement in MAE. Since MSE penalizes larger errors more heavily, this indicates that our imputed data are especially effective at helping the forecasting models avoid significant prediction mistakes. By providing a more complete and coherent history, the data imputed by PRAIM enables the models to better understand and project the underlying demand patterns, leading to more reliable and accurate forecasts.

\section{Conclusion and Future Work} \label{sec:conclusion}

In this paper, we address the critical and often overlooked challenge of missing data in real-world EV charging datasets. We identify the key limitations of existing methods, including the inability to effectively fuse multimodal context, vulnerability to data sparsity, and reliance on a restrictive one-model-per-station paradigm. To overcome these challenges, we develop PRAIM, a novel probabilistic variational imputation framework. Our approach features a pre-trained LLM as a universal encoder to generate rich, contextual embeddings from heterogeneous data. This is coupled with a RAG mechanism that dynamically draws on the entire dataset as an external knowledge base, enabling the training of a single, universal imputation model. Validated on four diverse, real-world datasets, PRAIM demonstrates consistently and significantly superior performance over a wide range of established benchmarks.

Our analysis further reveals that PRAIM's success stems from its fundamentally different approach to imputation. By leveraging an LLM, it goes beyond simple numerical relationships to capture the semantic context of each data point. RAG further empowers the model to overcome local data scarcity by finding and incorporating relevant precedents across the entire charging network, making it uniquely robust to high levels of missingness. Our experiments validate it and show that PRAIM not only achieves superior point-wise accuracy (more than $20\%$ MAE reduction) but also substantially preserves the statistical distribution of the original data (e.g., more than $25\%$ improvement in the Wasserstein distance), resulting in significant performance boosts in downstream forecasting tasks.

For future work, extending this LLM- and RAG-based imputation strategy to other complex time-series domains, such as urban mobility and electric demand, presents a compelling avenue for research.

\bibliographystyle{ieeetr}
\bibliography{IEEEabrv}

\begin{minipage}{.95\columnwidth}
    \begin{IEEEbiography}[{\includegraphics[width=1in,height=1in,clip,keepaspectratio]{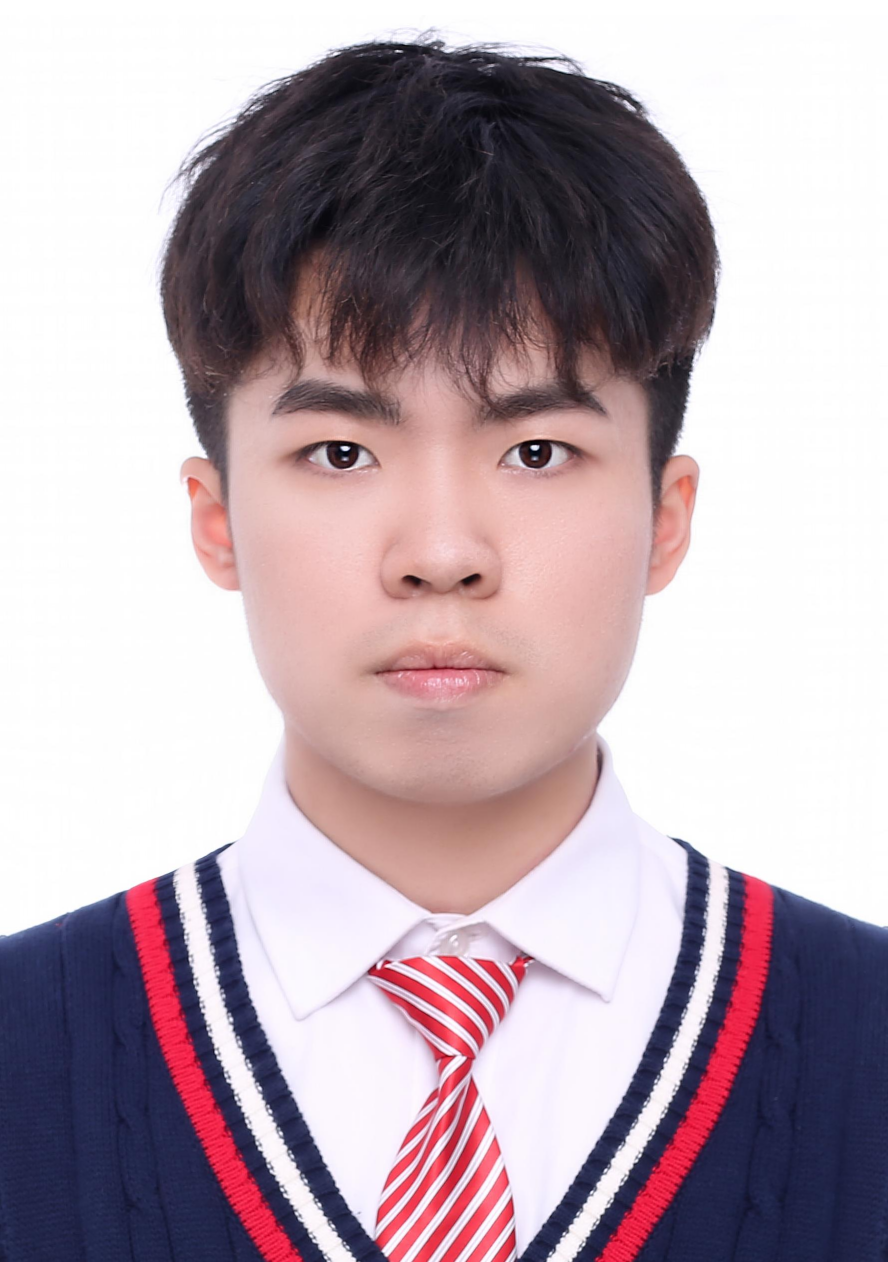}}]{Jinhao Li} received a B.E. degree in smart grid information engineering from the University of Electronic Science and Technology of China (UESTC) in 2022. He is currently pursuing his Ph.D. degree at the Department of Data Science and AI, Faculty of Information Technology, Monash University. His research interest is in machine learning for energy systems.
\end{IEEEbiography}
\end{minipage}

\begin{minipage}{.95\columnwidth}
    \begin{IEEEbiography}
[{\includegraphics[width=1in,height=1.25in,clip,keepaspectratio]{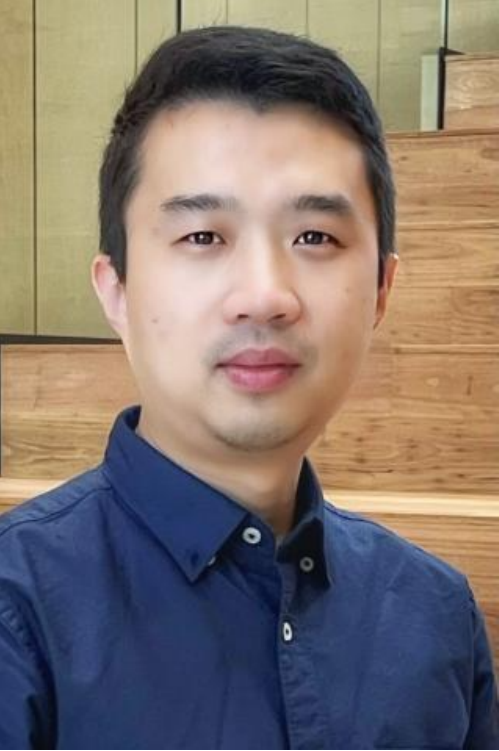}}]{Hao Wang} (M'16) received his Ph.D. in Information Engineering from The Chinese University of Hong Kong, Hong Kong, in 2016. He was a Postdoctoral Research Fellow at Stanford University, Stanford, CA, USA, and a Washington Research Foundation Innovation Fellow at the University of Washington, Seattle, WA, USA. He is currently a Senior Lecturer and ARC DECRA Fellow in the Department of Data Science and AI, Faculty of IT, Monash University, Melbourne, VIC, Australia. His research interests include optimization, machine learning, and data analytics for power and energy systems.
\end{IEEEbiography}
\end{minipage}

\section*{Appendix A. Theoretical Analysis of the Impact of Missing Data on Downstream Forecasting} 
\label{appendix:analysis_missing_data_impact}

Given the considerable amount of missing data shown in Fig. \ref{fig:missing_dist}, in the following, we analyze the impact of missing data on downstream tasks, by taking the common EV charging demand forecasting as an example. Missing values do not merely reduce the total volume of information; they can substantially diminish the number of usable samples for data-driven approaches particularly for forecasting \cite{wang2025survey}. Such reduction occurs since most forecasting models require a continuous, uninterrupted sequence of historical data as input.

For example, a DL-based forecasting model uses a look-back window of length $T$ to predict future charging demands with a horizon length of $H$. A single training sample consists of a sequence of $T+H$ data points. Even if one of these points is missing, the entire sequence becomes unusable for training.

Let $\delta$ be the proportion of missing daily records for a certain charging station, assumed to be randomly distributed. The probability that any given data point is available is thus $1-\delta$. The probability of deriving a complete training sample is thus $(1-\delta)^{T+H}$. For instance, to predict the next week's charging demand ($H=7$) using the past two weeks' data ($T=14$), we need a continuous sequence of $21$ days. In a dataset like Boulder, where the median proportion of missing data is about $20\%$, the probability of finding a single complete training sample is $(1-0.2)^{21}\approx 0.9\%$, demonstrating that even a moderate proportion of missing data can make it impossible to train forecasting models without prior imputation.

The problem is further compounded in spatiotemporal models, such as those using graph neural networks \cite{chen2022graph,kong2023graph}. These models leverage data not only from a target station but also from its $C$ neighbors from the constructed graph. The probability of a training sample being fully intact is $(1-\delta)^{(C+1)\times (T+H)}$, making the exponential decay now far more severe. Hence, the necessity of complete data across both temporal and spatial dimensions makes data imputation not just beneficial but a critical prerequisite for leveraging advanced forecasting models on EV charging demand.

\section*{Appendix B. Masking Fidelity Analysis} 
\label{appendix:masking_analysis}

In addition to the current random masking, to exactly reflect the shape of missingness, we develop a distribution-matched mask generator that reproduces the empirical structure of the actual missingness within the EV charging datasets. Specifically, for each charging station within each dataset, we estimate the contiguous gap-length distribution (e.g., one-day or two-day block), the day-of-week and seasonal frequencies of missingness, and the spacing distribution between gaps. Synthetic masks are then created by sampling gap blocks from the empirical gap-length probability mass function (PMF) and placing them at calendar positions drawn with the observed day-of-week/season weights, respecting sampled spacings and avoiding overlaps until the desired fraction of missing days is reached.

To assess whether our artificial masks (i.e., the $\lambda$ sweep and the distribution-matched mask generator) resemble real missingness, we use three complementary fidelity metrics. First, we compare the contiguous gap-length distributions (with lengths from one to six days) using the Wasserstein distance (WD) on a discrete support. 
Second, we compare the day-of-week (DoW) missingness profiles with Jensen–Shannon (JS) divergence, a symmetric, finite variant of KL divergence. 
Third, we examine short-range dependence in the missingness indicator via the difference in lag-1 autocorrelation (AC), denoted as $\Delta AC$, relative to the real series.

\begin{table*}[!t]
    \caption{Mask fidelity metrics (lower is better) comparing each synthetic masking regime to the empirical real missingness.}

    \centering

    \begin{tabular}{l|cc|cc|cc}

    \multirow{2}{*}{Dataset} & \multicolumn{2}{c|}{Gap length WD} & \multicolumn{2}{c|}{DoW JS divergence} & \multicolumn{2}{c}{Lag-1 $|\Delta AC | $} \\

     & DM vs Real & LS vs Real & DM vs Real & LS vs Real & DM vs Real & LS vs Real \\

    \hline
    
    Boulder    & \textbf{0.060} & 0.180 & \textbf{0.015} & 0.042 & \textbf{0.030} & 0.090 \\
    
    Dundee     & \textbf{0.050} & 0.140 & \textbf{0.012} & 0.035 & \textbf{0.025} & 0.070 \\
    
    Palo Alto  & \textbf{0.070} & 0.200 & \textbf{0.018} & 0.048 & \textbf{0.035} & 0.100 \\
    
    Perth      & \textbf{0.055} & 0.160 & \textbf{0.014} & 0.040 & \textbf{0.028} & 0.082 \\
    
    \hline
    
    \end{tabular}
    
    \label{tab:mask_fidelity_DM_vs_LS}    
\end{table*}

The results are provided in Table \ref{tab:mask_fidelity_DM_vs_LS}, where lower values represent a better imitation of the realistic missing patterns. The distribution-matched (DM) generator achieves consistently smaller WD than the $\lambda$-sweep (LS) for every dataset, indicating closer alignment with the empirical data missingness structure. This pattern reflects a core difference: DM explicitly samples blocks using the empirical gap-length PMF, whereas LS masks $\lfloor \lambda\cdot L \rfloor$ positions uniformly at random within a 7-day window, which naturally overproduces single-day holes and underproduces multi-day runs. The DoW JS divergence results align with the gap-length WD. DM tracks the mild weekday biases present in the logs since DM places blocks with observed calendar weights while LS is intentionally calendar-agnostic. As a result, the DoW profile derived from LS remains flatter and therefore more divergent from reality. Finally, the lag-1 $|\Delta AC |$ is uniformly smaller for DM (0.025–0.035) than LS (0.07–0.10), indicating the DM masks reproduce the clustering tendency of real outages, while LS looks more like independent draws within a window. These results clearly demonstrate that the DM faithfully mirrors both the rate and the structure of missingness, including blockiness, calendar effects, and short-range dependence, all of which can materially affect imputation difficulty. Using DM therefore provides a high-fidelity stress test that is closer to operational conditions. However, the values of LS are still significant even though it is structurally simpler than the DM. LS provides a conservative, bias-free baseline, by performing the masking uniformly at random within the $7$-day window, it avoids injecting any assumed calendar or spatial pattern into evaluation.

\section*{Appendix C. Probabilistic Imputation Performance Comparison} 
\label{appendix:prob_comparison}

Table \ref{tab:prob_comparison} shows that PRAIM delivers consistently stronger probabilistic performance than the best-performing LLM-excluded baseline for each dataset (MissForest for Boulder, matrix factorization for Dundee, GAN for Palo Alto, and denoising autoencoder for Perth). Specifically, PRAIM achieves lower NLL and CRPS in all four datasets (i.e., Boulder: 2.11/6.8 vs. 2.45/7.6; Dundee: 2.58/9.4 vs. 2.92/10.5). Since these metrics jointly assess accuracy of the mean and quality of the predictive spread, the gains indicate that PRAIM's distributions are not just sharper but also better aligned with realized outcomes. The advantage is largest where missingness and heterogeneity are more pronounced (e.g., Boulder), matching our design intuition that retrieval-grounded conditioning and a variational latent help when context is ambiguous.

\begin{table*}[!t]
    \caption{Probabilistic imputation performance comparison of PRAIM and the best-performing baseline across four datasets.}
    
    \centering
    
    \resizebox{2\columnwidth}{!}{
    \begin{tabular}{l|cccc|cccc|cccc|cccc}
    
    \multirow{2}{*}{Method} & \multicolumn{4}{c|}{Boulder} & \multicolumn{4}{c|}{Dundee} & \multicolumn{4}{c|}{Palo Alto} & \multicolumn{4}{c}{Perth} \\

     & NLL & CRPS & ICP@80 & ICP@95 & NLL & CRPS & ICP@80 & ICP@95 & NLL & CRPS & ICP@80 & ICP@95 & NLL & CRPS & ICP@80 & ICP@95 \\

    \hline
    
    PRAIM & \textbf{2.11} & \textbf{6.8} & \textbf{0.77} & \textbf{0.91} & \textbf{2.58} & \textbf{9.4} & \textbf{0.85} & \textbf{0.92} & \textbf{3.05} & \textbf{12.7} & \textbf{0.76} & \textbf{0.93}  & \textbf{2.34} & \textbf{8.6} & \textbf{0.75} & \textbf{0.90} \\

    Baseline & 2.45 & 7.6 & 0.69 & 0.84 & 2.92 & 10.5 & 0.71 & 0.85 & 3.38 & 14.1 & 0.70 & 0.83 & 2.61 & 9.7 & 0.67 & 0.82 \\

    \end{tabular}
    }

\label{tab:prob_comparison}
\end{table*}

\begin{figure}[!t]
    \centering
    \includegraphics[width=.9\linewidth]{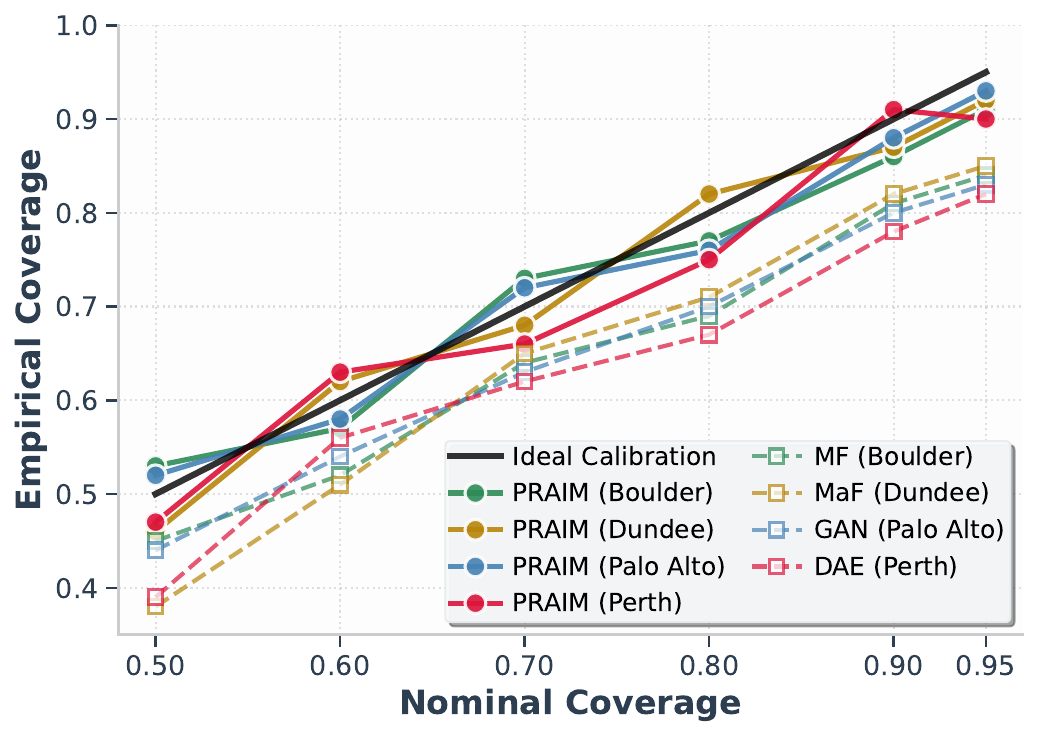}
    \caption{Calibration curves of PRAIM and best-performing baselines.}
    \label{fig:calibration_curves}
\end{figure}

In addition, calibration results align with the above findings. PRAIM's central prediction intervals achieve near-nominal coverage at both $80\%$ and $95\%$ levels (i.e., Boulder 0.77/0.91; Dundee 0.85/0.92; Palo Alto 0.76/0.93; Perth 0.75/0.90), whereas baselines under-cover by about $5$ to $8$ percentage points at each level. Such results suggest the baselines tend to produce intervals that are too narrow (i.e., over-confident), while PRAIM's variances are well-calibrated. In practical, PRAIM's intervals can be interpreted as reliable uncertainty bands for risk-aware decisions (e.g., flagging imputed values after long outages), rather than treating them as ad-hoc noise terms. Also, the calibration curves (when setting $\alpha$ as $\{0.5, 0.6, 0.7, 0.8, 0.9 ,0.95\}$) of both PRAIM and best-performing baselines are illustrated in Fig. \ref{fig:calibration_curves}.

\section*{Appendix D. Results of Replacing LLM Embeddings with Conventional Encoders} 
\label{appendix:conventional_encoder}

The results in Table \ref{tab:encoder_comp} show that, across all four datasets, the best conventional encoders (such as LSTM) substantially outperform purely tabular variants, but they still trail our PRAIM pipeline that uses LLM-derived context embeddings. The reason is structural: the LLM embedding (further grounded by RAG) fuses heterogeneous signals, including recent demand and its missingness, calendar rhythm, geospatial coordinates, and PoI semantics, into a single, compositional representation before decoding. This unified context lets PRAIM exploit feature interactions that lightweight temporal encoders only approximate with late fusion. Consistent with the per-dataset plots in Fig. \ref{fig:MAE_comparison}, PRAIM's curve sits below the strongest conventional choices across masking ratios, translating into lower MAE at the daily level while preserving distributional fidelity.

\begin{table}[!h]
    \caption{Imputation MAE when using conventional encoders.}
    \centering
    \begin{tabular}{c|c|c|c|c}
    
    \multicolumn{1}{c|}{\diagbox[height=2\normalbaselineskip]{Encoder}{Dataset}} & Boulder & Dundee & Palo Alto & Perth  \\

    \hline 
    
    Tabular MLP   & 33.4 & 38.6 & 54.2 & 36.9 \\

    \hline 
    
    Time2Vec      & 32.6 & 36.9 & 50.8 & 35.5 \\
    
    \hline 
    
    TCN           & 29.6  & 35.6 & 36.5 & 34.2 \\
    
    \hline 
    
    LSTM          & 29.8  & 33.1 & 34.7 & 28.3 \\
    
    \hline 
    
    Transformer   & 29.3  & 35.0 & 37.9 & 34.4 \\
    
    \hline 
    
    GCN           & 32.1 & 39.7 & 58.1 & 38.1 \\
    
    \hline 
    
    GAT           & 31.7 & 38.9 & 56.4 & 37.3 \\
    
    \hline 

    \end{tabular}
    \label{tab:encoder_comp}
\end{table}

\section*{Appendix E. Results of Using Advanced Prompt Engineering Strategies} 
\label{appendix:prompt_strategies}

We introduce prompt engineering strategies, including Schema-First, Slot-Filled (SFSF),  Few-Shot, Station-Typed (FSST), Decomposition-based Fusion (DF) \cite{R4C1_prompt_1,R4C1_prompt_2} compared to our current template (which is instruction-style and loosely structured as shown in Fig. \ref{fig:example_prompt}).

\begin{table}[!t]
    \caption{Relative change in MAE using different prompt engineering strategies compared to the current prompt.}
    
    \centering
    
    \resizebox{\columnwidth}{!}{
    \begin{tabular}{c|c|c|c|c|c}
    
    \diagbox[height=2\normalbaselineskip]{Prompt}{Dataset} & Boulder & Dundee & Palo Alto & Perth  \\
    
    \hline 

    SFSF & $-3.6\%$ & $-2.9\%$ & $-4.1\%$ & $-3.2\%$ \\
    
    \hline
    
    FSST & $-5.8\%$ & $-6.4\%$ & $-7.1\%$ & $-5.3\%$ \\
    
    \hline
    
    DF   & $-4.7\%$ & $-5.1\%$ & $-5.9\%$ & $-4.4\%$ \\

    \hline

    \end{tabular}
    }
    
    \label{tab:prompt_ablation}
\end{table}

Compared to our current prompt, the relative change in mean absolute error (MAE) is reported in Table \ref{tab:prompt_ablation}, where negative/positive values represent imputation performance improvement/degradation. The prompt ablation confirms that better-engineered prompts do help. All three strategies consistently reduce MAE relative to the currently adopted prompt. Among them, FSST produces the largest average gains (around 5–7\% across datasets), followed by DF (4–6\%), then SFSF (3–4\%). This ordering matches intuition: FSST that specifically encodes EV-specific regimes (workplace vs. residential, weekday vs. weekend, holiday and gap patterns) guides the LLM to form more useful context embeddings. DF regularizes how temporal rhythm and spatial role are combined. A schema-only prompt, e.g., SFSF, still helps by reducing output entropy, but without examples, its effect is smaller.

Notably, the improvements remain comfortably below $10\%$, indicating that PRAIM's effectiveness is not tightly constrained by prompt engineering. Using the reported MAE ranges, i.e., $[13,16]$ for Palo Alto, $[13,20]$ for Boulder, $[4,5]$ for Dundee, and $[11, 15]$ for Perth, the uplift brought by advanced prompt engineering techniques translates to practical but minor absolute changes. These are welcome gains, but they also show that the PRAIM's core, i.e., RAG grounding alongside variational conditioning with FiLM-modulated decoding, already provides robust performance, with prompt design acting as a second-order refinement rather than a brittle bottleneck.

\section*{Appendix F. Results of Using Different Filter Thresholds} 
\label{appendix:filter_thresholds}

We let $\tau$ denote the filter threshold and run a threshold sweep $\tau \in \{20\%, 30\%, 40\%, 50\%\}$. We re-train PRAIM as well as re-building the RAG corpus for each $\tau$. Table \ref{tab:filter_thershold} reports the relative MAE change compared to the default value of $35\%$, where negative/positive values represent imputation performance improvement/degradation.

\begin{table}[!t]
    \caption{Relative change in MAE using different filter thresholds across four datasets compared to $\tau=35\%$.}
    \centering
    \begin{tabular}{c|c|c|c|c}
    
    \diagbox[height=2\normalbaselineskip]{Dataset}{$\tau$} & $20\%$ & $30\%$ & $40\%$ & $50\%$  \\
    
    \hline
    
    Boulder   & $+5.6\%$ & $+3.6\%$ & $+11.3\%$ & $+29.2\%$  \\
    
    \hline
    
    Dundee     & $+8.7\%$ & $+0.8\%$ & $+9.1\%$ & $+15.1\%$ \\
    
    \hline
    
    Palo Alto  & $+10.1\%$& $+1.5\%$ & $+8.9\%$ & $+26.0\%$ \\
    
    \hline
    
    Perth      & $+9.5\%$ & $+1.0\%$ & $+11.3\%$ & $+15.4\%$ \\
    
    \hline
    
    \end{tabular}
    \label{tab:filter_thershold}
\end{table}

The results show an asymmetric U-shape around $\tau=35\%$. On the one hand, over-permissive thresholds, i.e., large $\tau$, allow stations with very high missingness inject structural noise (e.g., quasi-abandoned or intermittently disconnected units) and contaminates retrieval with low-evidence neighbors. Such thresholds sharply degrade imputation accuracy. For example, at $\tau=50\%$, MAE worsens by $29.2\%$ for Boulder, $26.0\%$ for Palo Alto, $15.1\%$ for Dundee, and $15.4\%$ for Perth. On the other hand, for over-strict thresholds, i.e., small $\tau$, excluding many stations that have moderate but usable history shrinks station diversity, narrows coverage of EV charging patterns, and thins the RAG corpus, thereby reducing the chance of finding strong analogues for difficult cases. Such thresholds produce smaller but consistent degradations. For example, at $\tau=20\%$, MAE increases by $5.6\%$ for Boulder, $10.1\%$ for Palo Alto, $8.7\%$ for Dundee, and $9.5\%$ for Perth. Setting $\tau=30\%$ is closer to the current optimum (i.e., $\tau=35\%$) but still above it in all four datasets. Thus, from an empirical perspective, setting the value of threshold of $35\%$ tend to filter out stations whose history is too unreliable while preserving enough cross-station heterogeneity to make RAG effective and training stable.

\end{document}